\documentclass[10pt,letterpaper]{article}
\usepackage[top=0.85in,left=2.75in,footskip=0.75in]{geometry}

\usepackage{amsmath,amssymb}

\usepackage{changepage}

\usepackage[utf8x]{inputenc}

\usepackage{textcomp,marvosym}

\usepackage{cite}

\usepackage[table]{xcolor}

\definecolor{linkcolor}{HTML}{005075}
\usepackage[colorlinks = true, linkcolor = linkcolor, urlcolor=linkcolor, citecolor = linkcolor]{hyperref}
\usepackage{nameref}

\usepackage[right]{lineno}

\usepackage{microtype}
\DisableLigatures[f]{encoding = *, family = * }

\usepackage{array}

\newcolumntype{+}{!{\vrule width 2pt}}

\newlength\savedwidth

\raggedright
\usepackage[aboveskip=1pt,labelfont=bf,labelsep=period,justification=raggedright,singlelinecheck=off]{caption}

\usepackage[numbers,sort,compress]{natbib}
\makeatletter
\renewcommand{\@biblabel}[1]{\quad#1.}
\makeatother

\usepackage{lastpage,fancyhdr,graphicx}
\usepackage{epstopdf}
\fancyheadoffset[L]{2.25in}
\fancyfootoffset[L]{2.25in}
\usepackage[bb=boondox]{mathalfa}
\usepackage{booktabs}
\usepackage{multicol}
\usepackage{subcaption}

\newcommand{\ie}{\emph{i.\@e.\@,~}}      
\newcommand{\eg}{\emph{e.\@g.\@,~}}

\newcommand{\ints}{\ensuremath{\mathbb{Z}}}
\newcommand{\reals}{\ensuremath{\mathbb{R}}}
\renewcommand{\d}{\ensuremath{\,\mathrm{d}}}

\definecolor{LightYellow}{rgb}{1,1,0.88}
\definecolor{LightRed}{rgb}{1,0.88,0.88}

\DeclareMathOperator{\1}{\mathbb{1}}

\DeclareMathOperator{\AUC}{AUC}

\makeatletter
\newcommand\Autoref[1]{\@first@ref#1,@}
\def\@throw@dot#1.#2@{#1}
\def\@set@refname#1{
  \edef\@tmp{\getrefbykeydefault{#1}{anchor}{}}%
  \xdef\@tmp{\expandafter\@throw@dot\@tmp.@}%
  \ltx@IfUndefined{\@tmp autorefnameplural}%
  {\def\@refname{\@nameuse{\@tmp autorefname}s}}%
  {\def\@refname{\@nameuse{\@tmp autorefnameplural}}}%
}
\def\@first@ref#1,#2{%
  \ifx#2@\autoref{#1}\let\@nextref\@gobble
  \else%
  \@set@refname{#1}
  \@refname~\ref{#1}
  \let\@nextref\@next@ref
  \fi%
  \@nextref#2%
}
\def\@next@ref#1,#2{%
  \ifx#2@ and~\ref{#1}\let\@nextref\@gobble
  \else, \ref{#1}
  \fi%
  \@nextref#2%
}
\makeatother

\begin{document}

\vspace*{0.2in}

\begin{flushleft}
{\Large
\textbf\newline{Learning from few examples: Classifying sex from retinal
images via deep learning} 
}
\newline
\\
Aaron Berk\textsuperscript{1*},
Gulcenur Ozturan\textsuperscript{2},
Parsa Delavari\textsuperscript{2}
David Maberley\textsuperscript{3},
\"Ozg\"ur Y{\i}lmaz\textsuperscript{4},
Ipek Oruc\textsuperscript{2}
\\
\bigskip
\textbf{1} Department of Mathematics \& Statistics, McGill University, Montr\'eal, Canada
\\
\textbf{2} Department of Ophthalmology and Visual Sciences, University of British Columbia, Vancouver, Canada
\\
\textbf{3} Department of Ophthalmology, University of Ottawa, Ottawa, Canada
\\
\textbf{4} Department of Mathematics, University of British Columbia, Vancouver, Canada
\\
\bigskip

%
%





* aaron.berk@mcgill.ca

\end{flushleft}


\section*{Abstract}

Deep learning (DL) techniques have seen tremendous interest in medical imaging,
particularly in the use of convolutional neural networks (CNNs) for the
development of automated diagnostic tools. The facility of its non-invasive
acquisition makes retinal fundus imaging particularly amenable to such automated
approaches. Recent work in the analysis of fundus images using CNNs relies on
access to massive datasets for training and validation, composed of hundreds of
thousands of images. However, data residency and data privacy restrictions
stymie the applicability of this approach in medical settings where patient
confidentiality is a mandate. Here, we showcase results for the performance of
DL on small datasets to classify patient sex from fundus images --- a trait
thought not to be present or quantifiable in fundus images until
recently. Specifically, we fine-tune a Resnet-152 model whose last layer has
been modified to a fully-connected layer for binary classification. We carried
out several experiments to assess performance in the small dataset context using
one private (DOVS) and one public (ODIR) data source. Our models, developed
using approximately $2\,500$ fundus images, achieved test AUC scores of up to
$0.72$ ($95\%$ {CI}: $[0.67, 0.77]$). This corresponds to a mere $25\%$ decrease
in performance despite a nearly 1000-fold decrease in the dataset size compared
to prior results in the literature. Our results show that binary classification,
even with a hard task such as sex categorization from retinal fundus images, is
possible with very small datasets. Our domain adaptation results show that
models trained with one distribution of images may generalize well to an
independent external source, as in the case of models trained on DOVS and tested
on {ODIR}. Our results also show that eliminating poor quality images may hamper
training of the CNN due to reducing the already small dataset size even
further. Nevertheless, using high quality images may be an important factor as
evidenced by superior generalizability of results in the domain adaptation
experiments. Finally, our work shows that ensembling is an important tool in
maximizing performance of deep CNNs in the context of small development
datasets.




%
%

\section{Introduction}
\label{sec:introduction}

A retinal fundus reveals important markers of a patient's health~\cite{abramoff2010retinal, Wagner2020TVST}. Retinal fundus photography (\emph{retinal
  imaging}), in conjunction with manual image interpretation by physicians, is a
widely accepted approach for screening patient health conditions such as:
referable diabetic retinopathy, diabetic macular aedema, age-related macular
degeneration, glaucoma, hypertension or
atherosclerosis~\cite{abramoff2010retinal}. Recent investigations into the
automated analysis of retinal images have suggested further improvements over
manual interpretation, whether via increased efficacy, efficiency or
consistency~\cite{abramoff2010retinal, choi2017multicateg,
  gulshan2016development, gulshan2019performance, krause2018grader,
  poplin2018prediction, raman2019fundus, ting2017development, sarunic2020_tvst}. In some cases,
automated image analysis has been able to detect patient attributes that were
previously thought not to be present or detectable in retinal images. For
example, a deep-learning based algorithm for automated retinal image analysis,
developed using an immense reference database of images, determined patient sex
from retinal images with impressive efficacy
($\AUC = 0.97$)~\cite{poplin2018prediction}.

Deep learning approaches to automated retinal image analysis are seeing increasing
popularity for their relative ease of implementation and high
efficacy~\cite{gulshan2016development, gulshan2019performance,
  poplin2018prediction, raman2019fundus}. Deep learning is a kind of machine
learning in which one or more neural networks, each having several
\emph{layers}, is \emph{trained} on a dataset so as to approximately minimize
its misfit between the model predictions and labels, typically using a
stochastic training procedure such as stochastic gradient
descent~\cite{bishop2006pattern, hardt2016train, murphy2012machine}. In the deep
learning setting, for a classification task, a neural network is a highly
overparameterized model (relative to the complexity of the \emph{learning task}
and/or dataset size) that has been constructed as a composition of alternately
affine and non-linear transformations (\ie layers). This neural network
\emph{architecture} is designed with the intent of modelling dataset features at
different scales, and several works have highlighted neural networks' efficacy
for modelling collections of natural images~\cite{goodfellow2014generative,
  he2016deep, heckel2018deepdecoder, lempitsky2018deep, radford2015unsupervised,
  simonyan2014very}.

In the work by~\citet{poplin2018prediction}, large deep learning models were
used to classify sex and other physiological and/or behavioural traits that were
linked to patient health based on retinal fundus images. The classifiers were
trained using Google computing services using a database of approximately 1.5
million images.  In~\citet{gulshan2016development}, an automated deep learning
algorithm was compared against manual grading by ophthalmologists for detecting
diabetic retinopathy (DR) in retinal images. The algorithm, based on the
Inception-V3 architecture~\cite{szegedy2016rethinking}, was trained on
approximately $128\,000$ images, and scored high sensitivity and specificity on
two independently collected validation sets. For example, they achieved a
(sensitivity, specificity) of $(90.3\%, 98.1\%)$ on one validation set when
tuning for high specificity; $(97.5\%, 93.4\%)$ on the same validation set when
tuning for high sensitivity. Results of~\citet{ting2017development} show
comparable efficacy, and demonstrate adaptivity of the algorithm to datasets
from multi-ethnic populations. The algorithm achieved an AUC of $0.879$ for
detecting referable DR.\@ The authors used an adapted VGGNet
network~\cite{simonyan2014very} as the model architecture. Another study trained
a deep learning CNN using over $130\,000$ color fundus images to detect of
age-related macular degeneration~\cite{burlina2017automated}. For further review
of deep learning applications to ophthalmology and retinal imaging
see~\cite{grewal2018deep,ting2019artificial,abramoff2010retinal,raman2019fundus,
  Wagner2020TVST}.

Despite their ease of use, deep neural networks can be notoriously difficult to
train~\cite{che2016mode, shalev2014understanding, srivastava2014dropout,
  srivastava2017veegan}. The high efficacy of automated approaches that was
touted above is typically achieved only with massive amounts of
data~\cite{gulshan2016development, poplin2018prediction}. For example, in a work
advertising its use of a ``small database'', \citet{choi2017multicateg} required
$10\,000$ images to achieve $30.5\%$ accuracy on a $10$-class classification
problem. Generally, it is unknown when deep learning models are able to achieve
high performance on small datasets. For example, theoretical sample 
complexity bounds guarantee generalization performance for certain neural
network architectures~\cite{bartlett2019nearly, shalev2014understanding}, but these
bounds are known to be loose~\cite{abu2012learning}. Likewise, the assumptions on the
training algorithm used to guarantee the generalization error bounds are
untenable due to non-convexity of the learning task. Thus, a relationship
between minimal dataset size and deep learning model performance is
uncharacterized. Moreover, it is possible this relationship cannot be
characterized in general due to the aforementioned non-convexity and \emph{intrinsic
  complexity} of a deep learning task on a particular dataset. In addition to
difficulties manifest in achieving high efficacy on a classification task on a
dataset using deep neural networks, there are further difficulties in
guaranteeing the \emph{stability} of the resultant trained
model~\cite{germain2020pac, lee2018minimax, shalev2014understanding}. For
example, if an automated method for image analysis was designed to achieve high
performance on a homogenous population of retinal images, how is its performance
affected when evaluated on a heterogeneous population of images?

Availability of very large, accurately labeled, high quality retinal imaging
datasets with associated meta-data is limited~\cite{khan2021global}. Access to
such existing datasets is further complicated by legal, ethical, technical, and
financial barriers. An open problem, crucial for widespread applicability,
democratization, and generalization of deep learning in medical imaging is
regarding whether high performance can be achieved with deep learning methods on
challenging image classification tasks when the image database is small. In this
study, we take a step toward examining this open question by developing and
evaluating a deep learning model to classify patient sex from retinal
images. The deep learning model architecture is a state-of-the-art
ResNet~\cite{he2016deep} architecture, whose weights have been pre-trained on
the ImageNet database~\cite{ILSVRC15}. The pre-trained network is fine-tuned for
classifying sex on a database of approximately $2\,500$ fundus
images. Furthermore, we investigate the stability of a neural network trained on
a small database of images. We approach this goal empirically through domain
adaptation experiments, in which the performance of deep learning models trained
on one dataset of retinal fundus images are then evaluated on another.


\section{Methods}
\label{sec:methods}

\subsection{Annotated retinal fundus image datasets}
\label{sec:methods-annot-img-db}

In this work we introduce two novel private datasets of retinal fundus images that have been
curated from the Vancouver General Hospital ophthalmic imaging department: labeled DOVS-i and DOVS-ii datasets for the remainder of this paper.

The DOVS-i dataset was created from a total of $2\,025$ images from $976$
patients. Specifically, it was created by randomly subsampling this collection
of images to include only a single pair (one left and one right) of retinal images from each patient, resulting
in a collection of $1\,706$ images from $853$ patients. The DOVS-ii dataset was
curated from a collection of $3\,627$ images from $1\,248$ patients. As with
DOVS-i, the DOVS-ii dataset was created by randomly subsampling the image
collection to include only a single pair of images from each patient: $2\,496$
images from $1\,248$ patients. For statistics regarding the make-up of each
dataset, refer to~\Autoref{tab:dovs-i-dset-stats, tab:dovs-ii-dset-stats},
respectively.

In this work we additionally use a publicly available dataset for which the
retinal images contained therein were annotated with patient sex
information. The so-called ODIR database of images~\cite{odir-database}
comprises $7\,000$ images from $3\,500$ patients. This database of images was
subsampled, by us, to create the ODIR-N and ODIR-C datasets. The ODIR-N dataset
is a subset of ODIR, containing only ``normal'' eyes (eyes without any annotated
abnormality). The ODIR-N dataset contains $3\,098$ images from $1\,959$
individuals. The ODIR-C dataset is a subset of the ODIR-N dataset, which was
subsampled by an ophthalmologist (Dr.\ Ozturan) to eliminate poor quality
images, as image quality may contribute to classifier performance
(\eg~\cite{fu2019evaluation}). Given there are no clinical features in a retinal
fundus image relevant to sex classification, quality assessment focused on
ensuring that, as in standard fundus photography, the macula is in the centre of
image, and the optic disc is located towards the nose, and that any extreme
imaging artifacts were absent. Images with poor focus and illumination were also
removed. The ODIR-C dataset contains $2\,577$ images from $1\,722$ patients. For
statistics regarding the make-up of each dataset, refer
to~\Autoref{tab:odir-n-dset-stats, tab:odir-c-dset-stats}, respectively;
see~\autoref{S1_data_statistics} for detailed statistics and criteria followed
for the image elimination process. Throughout this work, we may refer to the
size of a partition using the notation $n_{\text{partition}}$. For instance, the
size of the validation partition of the DOVS-i dataset is
$n_{\text{val}} = 214$.

\begin{table}[h]
  \centering
  \null\hfill
  \begin{subtable}[h]{.4\textwidth}
    \begin{tabular}{lrrrr}
      \toprule
      \multicolumn{5}{c}{DOVS-i} \\
      \midrule
                                 & train & val & test & total\\
      \midrule
                        images   &  1280 & 214 &  212 & 1706 \\
      $\hookrightarrow$ female   &   656 & 112 &  108 &  876 \\
      $\hookrightarrow$ male     &   624 & 102 &  104 &  830 \\
      \midrule
                        patients &   640 & 107 &  106 &  853 \\
      $\hookrightarrow$ female   &   328 &  56 &   54 &  438 \\
      $\hookrightarrow$ male     &   312 &  51 &   52 &  415 \\
      \bottomrule
    \end{tabular}
    \caption{DOVS-i}
    \label{tab:dovs-i-dset-stats}
  \end{subtable}
  \hfill
  \begin{subtable}[h]{.3\textwidth}
    \begin{tabular}{rrrr}
      \toprule
      \multicolumn{4}{c}{DOVS-ii} \\
      \midrule
      train & val & test & total\\
      \midrule
      1746 & 374 &  376 & 2496 \\
       888 & 190 &  192 & 1270 \\
       858 & 184 &  184 & 1226 \\
      \midrule
       873 & 187 &  188 & 1248 \\
       444 &  95 &   96 &  635 \\
       429 &  92 &   92 &  613 \\
      \bottomrule
    \end{tabular}
    \caption{DOVS-ii}
    \label{tab:dovs-ii-dset-stats}
  \end{subtable}
  \hfill\null

  \vphantom{M}

  \null\hfill
  \begin{subtable}[h]{.4\textwidth}
    \begin{tabular}[h]{lrrrr}
      \toprule
      \multicolumn{5}{c}{ODIR-N} \\
      \midrule
                                & train & val & test & total \\
      \midrule
      images                    & 2170  & 470 &  458 & 3098  \\
      $\hookrightarrow$ female  &  980  & 209 &  209 & 1398  \\
      $\hookrightarrow$ male    & 1190  & 261 &  249 & 1700  \\
      \midrule
      patients                  & 1371  & 294 &  294 & 1959  \\
      $\hookrightarrow$ female  &  631  & 135 &  135 &  901  \\
      $\hookrightarrow$ male    &  740  & 159 &  159 & 1058  \\
      \bottomrule
    \end{tabular}
    \caption{ODIR-N}
    \label{tab:odir-n-dset-stats}
  \end{subtable}
  \hfill
  \begin{subtable}[h]{.3\textwidth}
    \begin{tabular}[h]{rrrr}
      \toprule
      \multicolumn{4}{c}{ODIR-C} \\
      \midrule
      train & val & test & total \\
      \midrule
      1816  & 381 & 380  & 2577  \\
       829  & 170 & 173  & 1172  \\
       987  & 211 & 207  & 1405  \\
      \midrule
      1205  & 258 & 259  & 1722  \\
       553  & 118 & 119  &  790  \\
       652  & 140 & 140  &  932  \\
      \bottomrule
    \end{tabular}
    \caption{ODIR-C}
    \label{tab:odir-c-dset-stats}
  \end{subtable}
  \hfill\null
  \caption{Dataset statistics. Counts are given for the total number of images
    and patients, and for each of the train, validation and test sets. In
    addition, counts for the number of female and male images/patients in each
    of the datasets, and in aggregate, are included.}
  \label{tab:dataset-statistics}
\end{table}

\subsection{DOVS dataset curation and pre-processing}
\label{sec:data-curation}

The corpus of DOVS images is comprised of color retinal fundus photographs
obtained from the Vancouver General Hospital Ophthalmic Imaging
Department. Individuals were randomly sampled from this database --- without
regard to age, sex or health status. For each individual sampled from the
database, a single pair of images (left and right eye) was added to the data
set. The individuals were assigned a unique token; this token, the sex of the
individual and a left/right identifier were used to label each image. We
describe each dataset below; see \autoref{S1_data_statistics} for more detailed
data statistics.

\paragraph{DOVS-i dataset}

A total of $853$ patients were selected from the database, yielding a data set
of $1\,706$ images. The resolution of each retinal fundus image was
$2\,392 \times 2\,048$ pixels. Before training, each image was resized to
$300\times 256$ using the Haar wavelet transform. Specifically, let
$\mathcal{W}$ denote the Haar wavelet transform and let $x$ denote a retinal
fundus image. Let $w := \mathcal{W}x$ so that $w_{j}$ is a tensor of the $j$th
level of detail coefficients, where $j = 0$ corresponds to the coarsest level
scaling coefficients. We remove the finest 3 scales by projecting onto the first
$W - 3$ levels, and inverse-transform $\tilde w := (w_{j})_{j=0}^{W-3}$ via
$\tilde x = \mathcal{W}^{-1}\tilde w$ to obtain the appropriate down-sized
image.

Individuals were randomly partitioned into training, validation and test sets in
a way that maximally resembled the proportion of females to males in the
aggregate population. This also ensured that both left and right eyes of an individual were always retained within the same partition. The training set was comprised of $640$ patients (\ie
$\sim 75 \%$ of the data at $1280$ images), the validation set of $107$ patients
($\sim 12.5\%$ of the data at $214$ images), and the test set of $106$ patients
($\sim 12.5\%$ of the data at $212$ images). Counts for each partition,
including counts stratified by sex, may be found
in~\autoref{tab:dovs-i-dset-stats}.

\paragraph{DOVS-ii database}

The procedure used for the DOVS-ii dataset was similar to that for the DOVS-i
dataset. The main difference is in the counts --- of images and patients. A
total of $1\,248$ patients were selected from the database, yielding a dataset
of $2\,496$ images. The set of unique patients included in DOVS-ii is a superset
of DOVS-i, however, since image selection for each patient was based on random
sampling of multiple existing images for both DOVS-i and DOVS-ii, this is not
true for the two image sets. The original size of each image is consistent with
those for the DOVS-i dataset; each was again subsampled using the aforementioned
Haar wavelet transform. The partitioning of the dataset into training,
validation and test sets was $70\%/15\%/15\%$, retaining both left and right
fundus images of each individual within the same partition. Counts for each
partition, including counts stratified by sex, may be found
in~\autoref{tab:dovs-ii-dset-stats}.

\paragraph{Ethics statement}

This is a retrospective study of archived samples. The study was approved by the UBC Clinical Research Ethics Board (H16-03222) and Vancouver Coastal Health Research Institute (V16-03222), and requirement of consent was waived.

\subsection{Network architecture}
\label{sec:network-architecture}

The architecture of the deep learning model used in this work was modified from
a deep residual network, ResNet-152~\cite{he2016deep}. To our knowledge, the
application of this network architecture to automated retinal image analysis
tasks has not previously been explored, and neither for the classification of
patient sex. Leveraging the common deep learning tool of transfer
learning~\cite{ng2016nuts, pan2009survey, shao2014transfer, tan2018survey,
  zhuang2020comprehensive}, the weights of the network are initialized using the
original weights~\cite{he2016deep} learned for the ILSVRC'15 ImageNet image
classification contest~\cite{ILSVRC15}. The final layer of the network was
replaced by a new layer suitable for a binary classification task such as
classifying patient sex. In our implementation of the transfer learning
approach, during training the model weights are ``fine-tuned'', such that all
weights in the network are updated. Updates were performed using stochastic
gradient descent. For more details on training implementation and model
architecture, see \nameref{S1_methods} and \nameref{S1_parameter_values}.

\subsection{Network training}
\label{sec:network-training}

A ``fine-tuning''~\cite{zhou2017fine} approach was used to train the network. In
this framework, all weights of the network are allowed to vary, with the
intuition that important features from the pre-trained layers will vary little
during training, having already reached approximate convergence for a large
class of natural images. The expectation, then, is that the as yet untrained weights in
the final fully connected layers will learn good features using the layer
outputs from the rest of the network.

The criterion determining such a goodness of fit was chosen to be binary
cross-entropy (BCE) loss. Specifically, BCE loss measures how well the network
performs on the task of labeling the training data. Let $x$ denote a particular
data point and suppose $h(x)$ takes on values between 0 and 1 representing the
``belief'' that x belongs to class 0 or 1. For example, $h(x)=0$ (or $h(x)=1$)
would mean that we believe $x$ belongs to class 0 (or class 1) with certainty
whereas intermediate values correspond to uncertain beliefs. With this in mind,
the BCE loss $\ell(h(x), y)$ represents the level of ``surprise'', or
discrepancy, between the true label $y$ and $h(x)$, namely:
\begin{align*}
  \ell(h(x), y) := - y \log h(x) - (1-y) \log (1 - h(x)). 
\end{align*}
Observe that $y \in \{0, 1\}$ and so exactly one of the above terms is non-zero
for any image-label pair $(x, y)$ in the database. Furthermore, by convention we set $0\log 0=0$.

The network was trained for a maximum of $300$ passes through the training data,
so-called epochs. On each epoch, the training algorithm was fed consecutive
batches of the training data, from which approximate gradients of the loss
function were computed to update the weights of the network. A batch size of 8
was used. 

\begin{figure}[h]
  \centering
  \includegraphics[width=.7\textwidth]{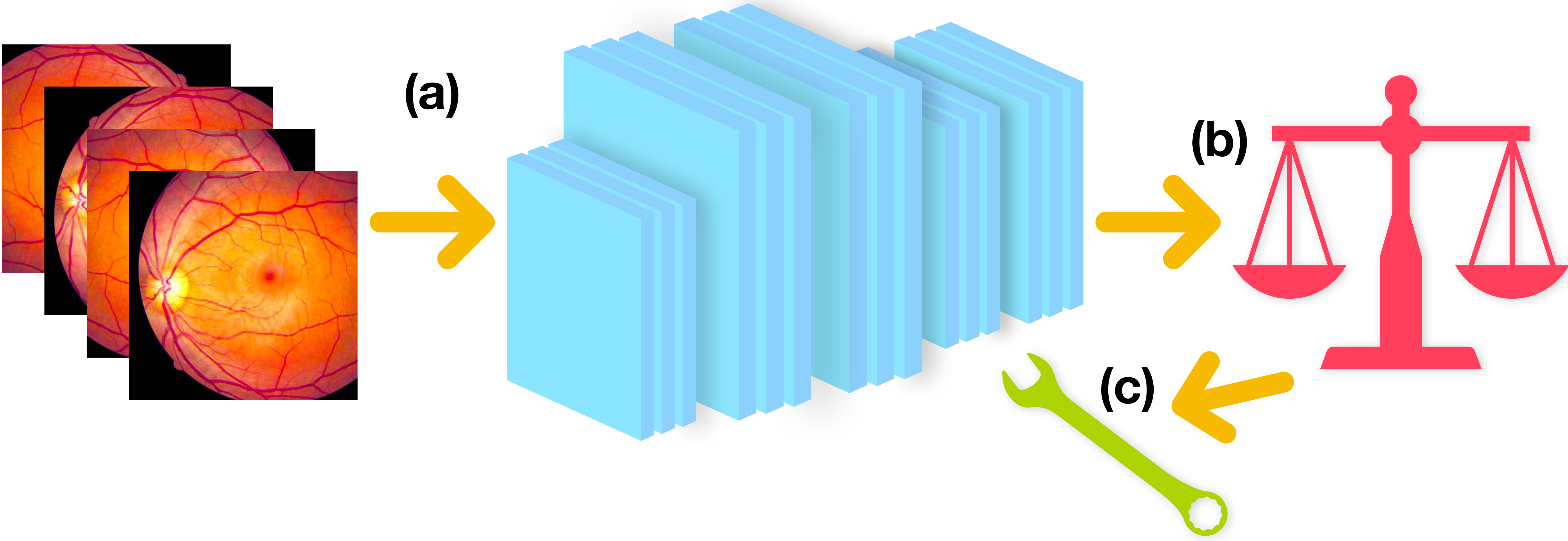}
  \caption{Training method: (a) a batch of images are fed into the network; (b)
    the batch loss is computed from the labels and model's output; (c) the
    optimizer updates the network's parameters via back-propagation. This
    procedure is repeated for many epochs until convergence is reached.}
  \label{fig:nn-method}
\end{figure}

For each epoch, the order of the images was randomized. An image is loaded only
once per epoch. For each image in each batch in each epoch, image augmentation
was performed such that the image was randomly rotated with probability $0.2$ by
a random angle between $-10$ and $10$ degrees inclusive; randomly cropped to a
size of $224 \times 224$ pixels; a horizontal flip was applied with probability
$0.3$; histogram equalization was applied; and then the image was normalized
using the standard mean and standard deviation parameters expected by ResNet:
\begin{align*}
  \mu_{(r, g, b)} &= [0.485, 0.456, 0.406] & \sigma_{(r, g, b)} &= [0.229, 0.224, 0.225]. 
\end{align*}
A graphical depiction of the network training procedure is given
in~\autoref{fig:nn-method}.

The network was validated after each epoch on the validation set
to determine both the validation loss and accuracy. Early stopping was employed
to mitigate overfitting of the model during training. Specifically, model
training was run for at least 3 epochs and effectively halted after validation
accuracy ceased to improve. Early stopping then selected the final model for
that model run to be the one from the epoch with highest-attained validation
accuracy. For background on early stopping, see~\cite{prechelt1998early}
or~\cite[Chapter~11.5.2]{hastie2009elements}. Properties of the validation sets
within each database are available in~\autoref{tab:dataset-statistics}.

During a model run, several metrics were recorded to track the progress of the
training procedure. In particular, accuracy (proportion correct), BCE and area
under the curve of the Receiver Operating Characteristic ($\AUC$) were measured
after each epoch for the model on both the training and validation partitions of
a dataset. A more detailed description of these three metrics may be found
in~\nameref{S1_methods}.

For each of the databases DOVS-i, DOVS-ii, ODIR-N, and ODIR-C, several model runs
were performed --- that is, several models were trained using the above
described training procedure. In particular, 5 models were trained using the
DOVS-i database; 6 each for the ODIR-N and ODIR-C databases. The models trained
using the DOVS-ii database were used for a model ensembling experiment. A
description of the development procedure for those models is reserved
to~\nameref{sec:ensembling-method}. Each trained model is labelled with an
identifier: D1--5 for the models trained using the DOVS-i dataset, E1--20 for
the models trained using the DOVS-ii dataset, N1--6 for the models trained
using the ODIR-N dataset, and C1--6 for the models trained using the ODIR-C
dataset. For sake of concision, we may refer to a model run metonymously with
the identifier associated to the trained model it produced (\eg ``model run
D1''). We disambiguate the two only where we feel it could otherwise lead to
confusion.

\subsection{Network evaluation}
\label{sec:network-evaluation}

The performance of a trained network was evaluated by computing a quantity of
interest on a ``hold-out'' set of data. In most cases, this hold-out set is
simply the test partition from the same dataset whose train and validation
partitions were used to train the model, and the quantity of interest is the
AUC score of the model. These are the results presented
in~\nameref{sec:results-prototypical}, and whose procedure is described
here. In~\nameref{sec:results-domain-adaptation}, where we present results of
domain adaptation experiments, the dataset used for test-time evaluation is no
longer the associated test partition, but an external dataset. A description of that procedure appears
in~\nameref{sec:domain-adaptation-method}.

For each test of a model from a run, the quantity of interest was that model's
area under the receiving operator characteristic (AUC) on the test set. For
more information on the receiver operating characteristic (ROC) and AUC, see
\nameref{S1_methods}. Significance of the results was evaluated using
bootstrapping, as described in \nameref{S1_bootstrap_method}.  A
$(1-\alpha)$-confidence interval was computed from $B = 1000$ bootstrap
replicates where $\alpha = 0.05$; $p$-values for each test were computed and
compared with an $\alpha = 0.05$ significance level after being adjusted for
multiple comparisons using the Benjamini-Hochberg procedure
(BH)~\cite{efron1994introduction}. Namely, one bulk adjustment for multiple
comparisons was performed, including all significance tests performed in this
work. As such, this adjustment includes tests for models D1--5, N1--6 and C1--6
presented in~\nameref{sec:results-prototypical}
(\emph{cf.}~\autoref{tab:confidence-intervals}). Other tests included in the
adjustment will be described in the appropriate sections. Throughout, a
significance level of $\alpha = 0.05$ was used. To each trained model is
associated two tests: its AUC score on the validation set, and that on the
test set. In tests for significance, the AUC score was compared to a reference
score corresponding to random chance: $\mu^{\text{ref}}_{\AUC} := 0.5$.

It is worth noting that tests for significance of a model's AUC score on the
training set were not performed. Models were only selected if they converged;
effectively, if a model converged, its AUC score on the training set is expected
to be large by construction. At any rate, the desideratum of the present work is
a model's performance on out-of-sample data, which would be the key indicator of
its performance in a clinical setting.

\subsection{Domain adaptation method}
\label{sec:domain-adaptation-method}

In this experiment, we investigated the efficacy of our approach for
\emph{transductive} transfer learning~\cite{joachims1999transductive}, in which
the distribution of retinal images underlying the test set is different from
that for the training set. The ability of a trained model to perform domain
adaptation in this manner provides an empirical understanding of the network's
\emph{stability} on the learning task: how its performance varies when the
underlying data distribution is varied. The experimental set-up was similar to
that described in \nameref{sec:network-training} and
\nameref{sec:network-evaluation}. The key difference is that models developed
using the training and validation partitions of one dataset were tested on
another \emph{entire} database. Namely, models D1--5 were developed using the
DOVS-i training and validation partitions and tested on both the ODIR-N and
ODIR-C databases; models E1--10 and E$^{*}$ were developed using the DOVS-ii
training and validation partitions and tested on both the ODIR-N and ODIR-C
databases; models N1--6 were trained using the training and validation
partitions of the ODIR-N database and tested on the DOVS-ii database; models
C1--6 were trained using the training and validation partitions of the ODIR-C
database and tested on the DOVS-ii database. Results of these experiments may be
found in \nameref{sec:results-domain-adaptation}, reported
in~\autoref{tab:cross-test-confidence-stats}. As above, the quantity of interest
was the AUC score of the model on the testing database. Significance of the
results was evaluated using bootstrapping, and each collection of 22 tests
(see~\autoref{tab:cross-test-Di},~\ref{tab:cross-test-Ni},~\ref{tab:cross-test-Ci};
\autoref{tab:cross-test-Dii}, respectively) was adjusted for multiple
comparisons using BH, as described in~\nameref{sec:network-evaluation}. As
above, $(1 - \alpha)$-confidence intervals were computed using bootstrapping
techniques as described in~\nameref{S1_bootstrap_method}.

\subsection{Ensembling method}
\label{sec:ensembling-method}

In \nameref{sec:network-architecture} we describe how the models examined in
this work are deep neural networks with a particular architecture, which map
images to binary labels corresponding to the sex of the
corresponding patient. In this section, we describe an additional experiment, in
which a collection of such models are combined to form a single
meta-classifier. The aim of these experiments is to understand the impact on
classification performance of combining several stochastically trained
models. For a review of ensembling, see~\cite{shalev2014understanding, hastie2009elements}.

Fix numbers $\ell, L \in \ints$ satisfying $1 \leq \ell \leq L$. Fix a
collection of trained models $f_{i}, i = 1,\ldots,L$, which take a retinal image
as input, and output class probabilities (\eg the probability that the sex of
the patient corresponding to the image is ``male''). Without loss of generality
assume that the models are ordered according to their validation AUC scores
such that $\AUC(f_{1}) \geq \AUC(f_{2}) \geq \ldots \geq \AUC(f_{L})$. We
define the $(\ell, L)$-ensemble of a collection of models as the classifier
whose output probabilities are the average of the output probabilities of the
best $\ell$ models:
\begin{align*}
  f^{*}(x) := \ell^{-1} \sum_{i = 1}^{\ell} f_{i}(x). 
\end{align*}

In this work, we investigate the performance of a $(10, 20)$-ensemble classifier
constructed from ResNet-152 models that were developed using the DOVS-ii
dataset. These models correspond with $E1-20$, which were introduced
in~\nameref{sec:network-training}. Efficacy of the ensemble was evaluated using
the AUC score of the ensemble on the test partition of the DOVS-ii
database. Significance of the scores (for the individual models and the
ensemble) was determined using bootstrapping techniques described in
\nameref{S1_bootstrap_method}, including the determination of confidence
intervals for the test AUC score at an $\alpha = 0.05$ confidence level. The
AUC scores were compared to the reference null mean for AUC of
$\mu^{\text{ref}}_{\AUC} = 0.5$. As described previously, we adjusted
for the multiple comparisons presented in this work using the BH
procedure. Specifically, the $11 \times 2$ significance tests related to the
model ensembling experiment (\emph{cf.}~\autoref{tab:conf-stats-ensemble}) were
included in the bulk adjustment described in~\nameref{sec:network-evaluation}.


\section{Results}
\label{sec:results-main}

\subsection{Training metrics}
\label{sec:training-metrics}

\begin{figure}[h]
  \centering
  \includegraphics[width=\textwidth]{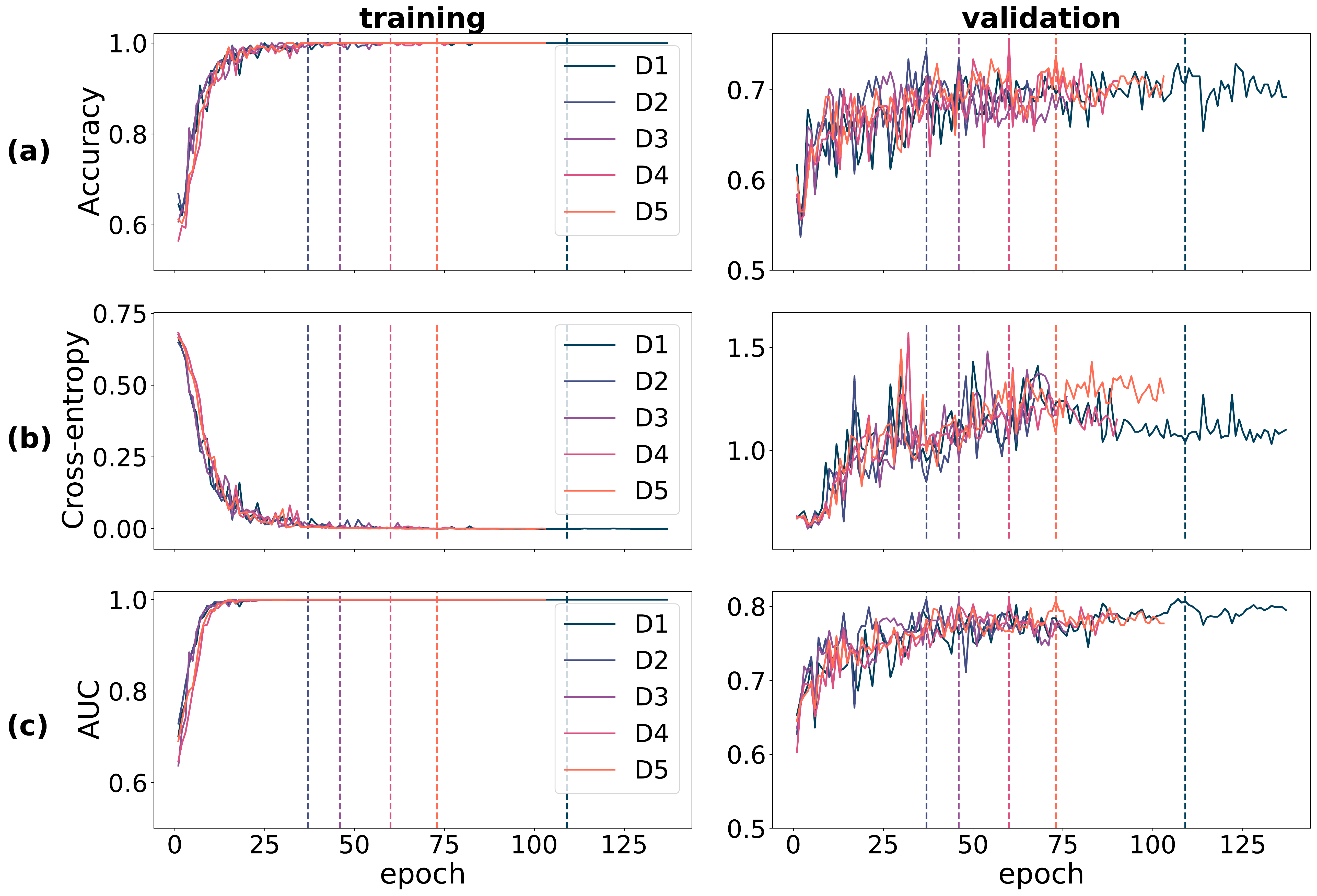}
  \caption{Training-time metrics for runs D1 through D5. (a) accuracy score (proportion correct);
    (b) binary cross-entropy loss; (c) AUC score. Vertical dashed lines correspond with the best epoch as selected by early stopping.}
  \label{fig:dovs-i-training-metrics}
\end{figure}

\begin{figure}[h]
  \centering
  \includegraphics[width=\textwidth]{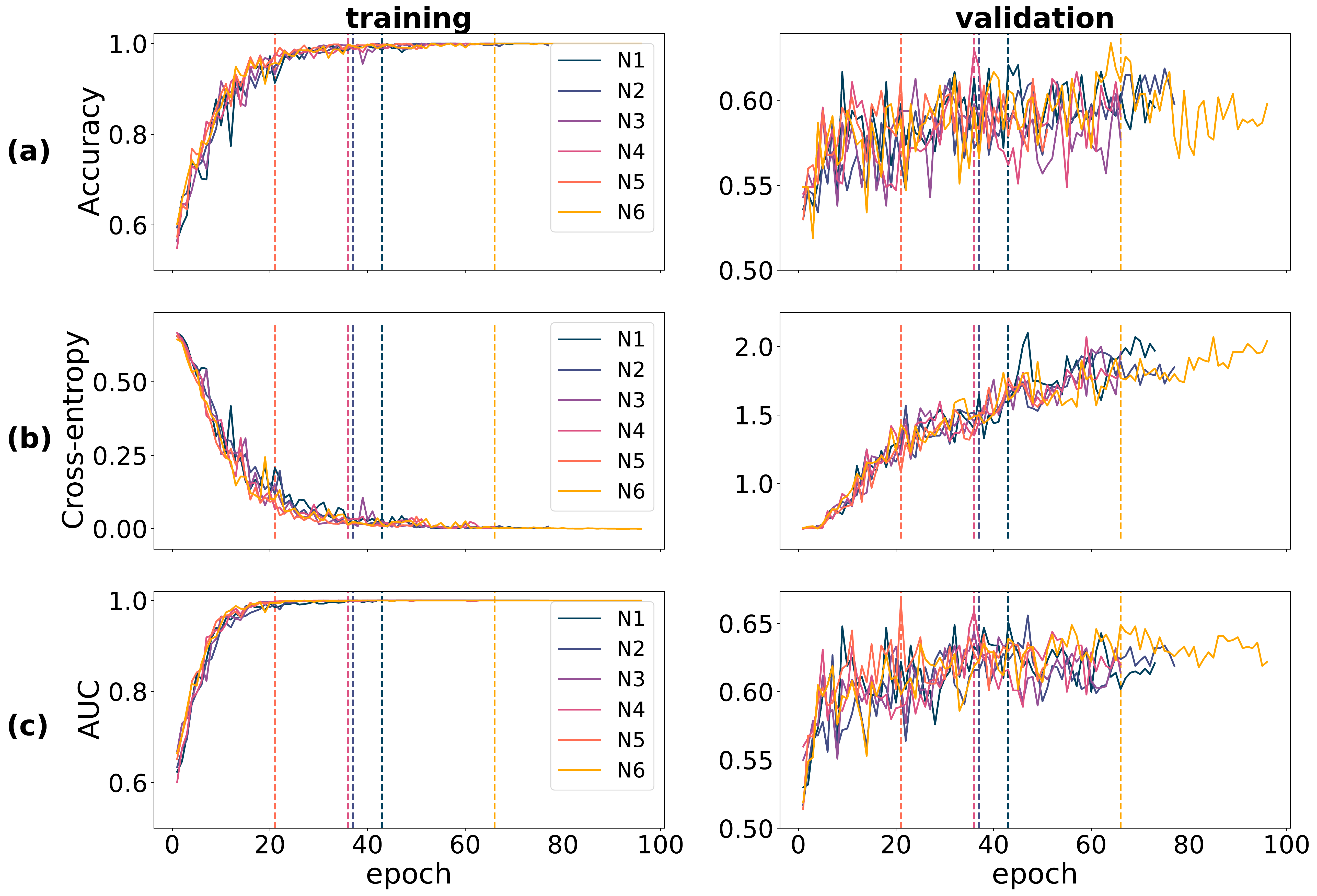}
  \caption{Training-time metrics for runs N1 through N6. (a) accuracy score (proportion correct);
    (b) binary cross-entropy loss; (c) AUC score. Vertical dashed lines correspond with the best epoch as selected by early stopping. }
  \label{fig:odir-n-training-metrics}
\end{figure}

\begin{figure}[h]
  \centering
  \includegraphics[width=\textwidth]{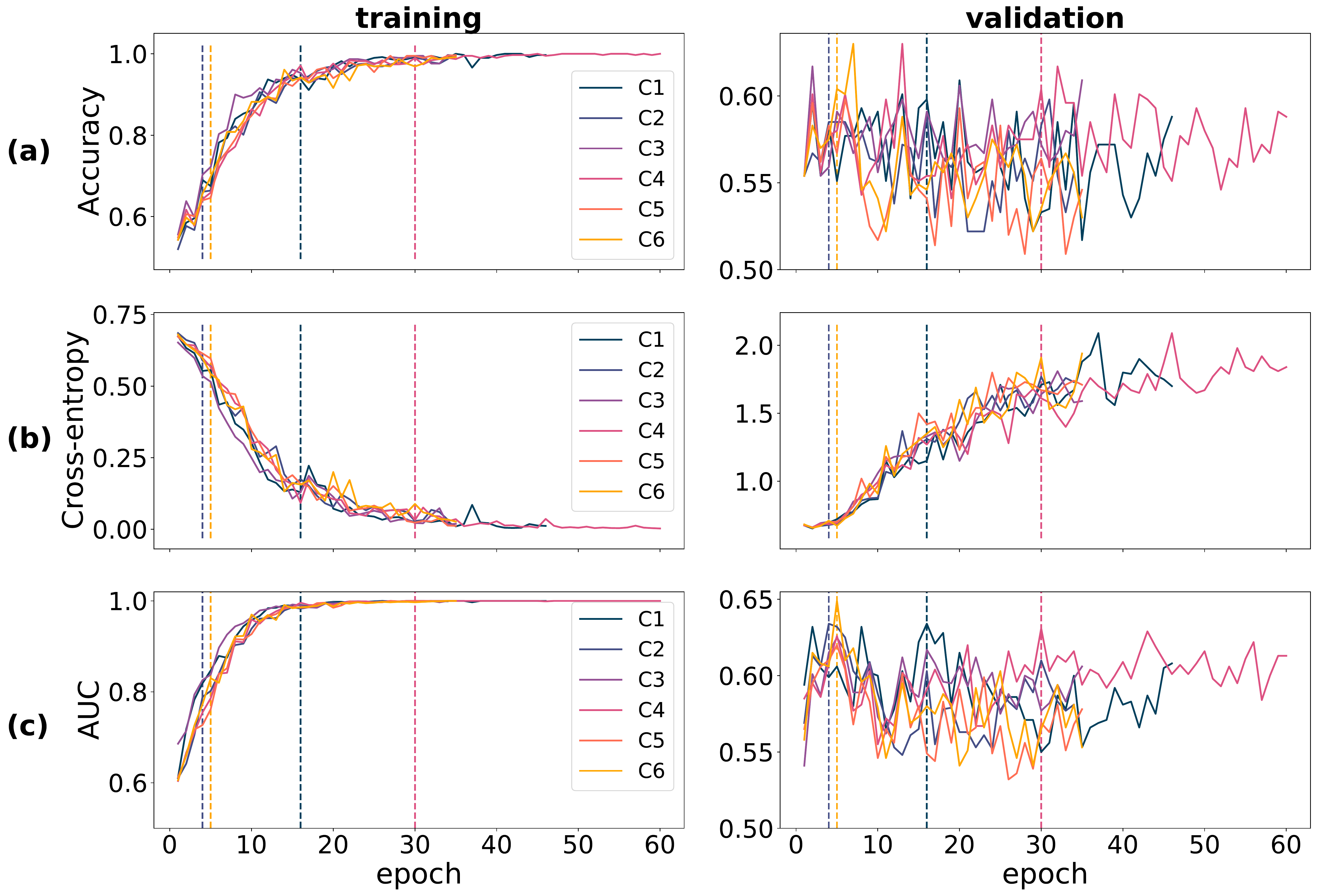}
  \caption{Training-time metrics for runs C1 through C6. (a) accuracy score (proportion correct);
    (b) binary cross-entropy loss; (c) AUC score. Vertical dashed lines correspond with the best epoch as selected by early stopping. }
  \label{fig:odir-c-training-metrics}
\end{figure}

Training-time metrics on both the train and validation partitions track a
model's performance during training, and determine the extent of success of the
training procedure. For all three datasets used in training --- DOVS-i, ODIR-N
and ODIR-C --- accuracy, cross-entropy loss and AUC score were recorded for
each training epoch. The data corresponding to these three metrics were plotted
graphically as a function of model
epoch. In~\autoref{fig:dovs-i-training-metrics}, a collection of six plots
stratifies the data for D1 through D5 (pertaining to DOVS-i) according to partition (training/validation) and each of the
three metrics. Specifically, accuracy appears in row (a), BCE in (b) and AUC
in (c). Each line within an individual plot corresponds to the data collected
for a single model run. The five DOVS-i model runs (D1--5) were plotted together to ease
the graphical interpretation of aggregate trends during training. Similar data
for training time metrics pertaining to model runs using the ODIR-N data are
depicted in~\autoref{fig:odir-n-training-metrics}; ODIR-C data,
in~\autoref{fig:odir-c-training-metrics}. For each model run, the final epoch
with plotted data varies due to the early stopping condition that halted model
training. The dashed vertical lines in each figure correspond with that epoch's
model weights that were selected as the final model for that model run. The
vertical lines correspond with the plotted data according to the colour, as
depicted in the legend.

In our experiments, we observed that convergence of the training accuracy
corresponded with that of the training AUC;\@ inversely so with the training
cross-entropy loss. Further, we observed that the validation accuracy and AUC
corresponded with their training set analogues, albeit with more noise and
overall lower scores, as should be expected.\footnote{By ``noise'', here, we
  mean the variance of the training metric as a function of the run epoch.} In
contrast, we observed that the cross-entropy loss behaved relatively erratically
on the validation set. Notably, unlike that for the training partition, the
validation cross-entropy loss \emph{increased} for a majority of the epochs
before stabilizing. This pattern is not inconsistent with the validation of a
complex model trained on a small dataset. We conjectured that this behaviour
comes about due to two phenomena: (1) the model's tendency to move towards more
confident responses as training progresses, resulting in a few incorrect but
confident responses coupled with (2) the imbalanced nature of BCE loss that
imposes unbounded penalty to confident incorrect decisions. More specifically,
the drop in the loss from one epoch to another when an uncertain incorrect
decision is replaced with a a confident correct decision is much smaller
compared to the increase in the loss when an uncertain incorrect decision is
replaced with a confident incorrect decision. To test this hypothesis, we
carried out an experiment in which we replaced BCE loss with a ``balanced'' loss
function. Consistent with our expectation, we found that validation loss
decreased in tandem with training loss when the loss function is balanced (see
\nameref{S1_loss} for methodological details and results of this experiment). It
is worth mentioning here that although this supplementary experiment allows us
to describe the conditions that bring about an increasing validation loss
concurrently with improvements in other performance metrics, it does not explain
why the model produces these few confident incorrect decisions. Such an account
is beyond the scope of the present study.

For models trained using the DOVS-i data, convergence of the three metrics was
typically attained within the first 25--75 epochs. Atypical in this instance is
D1, which was obtained at epoch $109$. We observe that convergence on the
training data is attained faster than on the validation data; and that
performance on the validation data may continue to improve even after the scores
on the training data are near-perfect. This observation is consistent with
modern deep learning research in which training typically proceeds well after
$0$ training loss is attained.

For model runs training using the ODIR-N data, 
as with D1--5, the training metrics for N1--6 generally
corresponded with another, except in the case of of validation BCE.\@ We
observed that validation accuracy was relatively noisier for N1--6 than for
D1--5. For all six runs, model convergence was reached within 20--70 epochs, as
depicted by the vertical dashed lines in each panel
of~\autoref{fig:odir-n-training-metrics}. Again we observe that convergence on
the training data is attained more quickly than on the validation data; and that
performance on the validation data may continue to improve even after the scores
on the training data are near-perfect.

For model runs training using the ODIR-C data and yielding models C1--6, 
we observe
that both validation accuracy and AUC appear to be markedly more erratic, compared to N1--6, as a
function of model epoch. Given that ODIR-N already has few images relative to the complexity of the model architecture, it is unsurprising to observe more erratic and lower validation scores from training with ODIR-C, which is an even smaller dataset of images. Interestingly, all six models (C1--6) were selected by early
stopping from within the first 30 epochs. This is indicative that the training
procedure may be prone to overfitting on this especially small dataset. We
reserve a more detailed comparison between DOVS-i and ODIR-C to
\nameref{sec:discussion-model-db-comparison}.

\subsection{Single-domain results}
\label{sec:results-prototypical}

In the single-domain machine learning task, a trained and validated model is
evaluated on an independent set of ``test'' data whose makeup resembles that of
the training data. In this section we include scores, confidence intervals and
significance values pertaining to experiments using a single database. We have
computed these values for both the validation and test sets so that the
validation performance may serve as a reference point. These values are
reported in~\autoref{tab:confidence-intervals}. Specifically, for models
trained on the DOVS-i training set and evaluated on the DOVS-i validation and
test sets, see~\autoref{tab:dovs-i-confidence-intervals}; for models trained on
the ODIR-N training set and evaluated on the ODIR-N validation and test sets,
see~\autoref{tab:odir-n-confidence-intervals}; and for models trained on the
ODIR-C training set and evaluated on the ODIR-C validation and test sets,
see~\autoref{tab:odir-c-confidence-intervals}. Finally, results for models
trained on the DOVS-ii training set and evaluated on the DOVS-ii validation and
test sets appear later in~\autoref{tab:conf-stats-ensemble}, as these models
were used in the ensembling experiment whose results are presented
in~\nameref{sec:results-naive-ensembling}.

The median test AUC score for the models D1--5 was $0.68$. The validation AUC
scores were all approximately $0.81$, suggesting overfitting to the validation
partition. The adjusted $p$-values $p_{\text{adj}}$ were computed using
$B = 1000$ bootstrap replicates with sample sizes $n_{\text{val}} = 214$ and
$n_{\text{test}} = 212$. For each of the tests for significance,
$p_{\text{adj}}$ was found to satisfy $p_{\text{adj}} \leq 1.1 \cdot 10^{-3}$,
implying the performance was significantly better than chance ($0.5$) at an
$\alpha = 0.05$ confidence level. \autoref{tab:dovs-i-confidence-intervals}
shows $95\%$ confidence intervals ($CI$) for the validation and test AUC scores
for each of the five models (D1--5).

The median test AUC score for models N1--6 was $0.62$, and the median validation
AUC score was $0.66$, suggesting that the extent of overfit was minimal. The
adjusted $p$-values $p_{\text{adj}}$ were computed using $B = 1000$ bootstrap
replicates with sample sizes $n_{\text{val}} = 470$ and $n_{\text{test}} =
458$. For each of the tests of significance, $p_{\text{adj}}$ was found to
satisfy $p_{\text{adj}} \leq 1.1 \cdot 10^{-3}$, implying the performance was
significantly better than chance ($0.5$) at an $\alpha = 0.05$ confidence
level. Confidence intervals for the test AUC score for these models at the same
confidence level were typically about $(0.57, 0.67)$, all six nearly
identical. All values for each of the six models (N1--6) are given
in~\autoref{tab:dovs-i-confidence-intervals}.

The median test AUC score for the models C1--6 was $0.59$, and the median
validation AUC score was $0.63$, suggesting the extent of overfit was
minimal. The adjusted $p$-values $p_{\text{adj}}$ were computed using $B = 1000$
bootstrap replicates with sample sizes $n_{\text{val}} = 381$ and
$n_{\text{test}} = 380$. For all tests for significance on the validation AUC,
$p_{\text{adj}}$ was found to satisfy $p_{\text{adj}} \leq 1.1 \cdot 10^{-3}$,
implying the validation scores were found to be significant at an
$\alpha = 0.05$ confidence level. For all but one test for significance on the
test AUC scores, $p_{\text{adj}}$ was found to satisfy
$p_{\text{adj}} \leq 7.1 \cdot 10^{-3}$, implying all but one of the test scores
were found to be significant at an $\alpha = 0.05$ confidence level. All values,
including $95\%$ confidence intervals, for each of the six models (C1--6) are
given in~\autoref{tab:dovs-i-confidence-intervals}.

\begin{table}[h]
  \begin{adjustwidth}{-2.25in}{0in}
    \centering
    \begin{subtable}[h]{.45\linewidth}
      \centering
      \begin{tabular}{lrrrcc}
      \toprule
      run & epoch & \multicolumn{2}{c}{$\AUC$} & \multicolumn{2}{c}{$\mathrm{CI}_{\alpha}$} \\
          &       &      val & test & val & test \\
      \midrule
      D1 & $109$ & $0.81^*$ & $0.69^*$ & $(0.75, 0.86)$ & $(0.62, 0.76)$ \\
      D2 &  $37$ & $0.81^*$ & $0.61^*$ & $(0.76, 0.86)$ & $(0.54, 0.68)$ \\
      D3 &  $46$ & $0.81^*$ & $0.69^*$ & $(0.74, 0.86)$ & $(0.62, 0.76)$ \\
      D4 &  $60$ & $0.81^*$ & $0.68^*$ & $(0.75, 0.86)$ & $(0.61, 0.75)$ \\
      D5 &  $73$ & $0.81^*$ & $0.67^*$ & $(0.75, 0.86)$ & $(0.59, 0.74)$ \\
        \bottomrule
      \end{tabular}
      \caption{\centering DOVS-i;\@ $n_{\text{val}} = 214, n_{\text{test}} = 212$.}
      \label{tab:dovs-i-confidence-intervals}
    \end{subtable}
    \qquad\qquad
    \begin{subtable}[h]{.45\linewidth}
      \centering
      \begin{tabular}{lrrrcc}
      \toprule
      run & epoch & \multicolumn{2}{c}{$\AUC$} & \multicolumn{2}{c}{$\mathrm{CI}_{\alpha}$} \\
          &       &         val &        test &           val &           test   \\
      \midrule
      N1 & $43$  & $0.65^*$ & $0.62^*$ & $(0.60, 0.70)$ & $(0.57, 0.67)$ \\
      N2 & $47$  & $0.66^*$ & $0.61^*$ & $(0.61, 0.70)$ & $(0.55, 0.66)$ \\
      N3 & $36$  & $0.64^*$ & $0.62^*$ & $(0.60, 0.69)$ & $(0.57, 0.66)$ \\
      N4 & $36$  & $0.66^*$ & $0.61^*$ & $(0.61, 0.70)$ & $(0.57, 0.67)$ \\
      N5 & $21$  & $0.67^*$ & $0.62^*$ & $(0.61, 0.72)$ & $(0.57, 0.67)$ \\
      N6 & $66$  & $0.65^*$ & $0.62^*$ & $(0.60, 0.70)$ & $(0.57, 0.68)$ \\
      \bottomrule
      \end{tabular}
      \caption{\centering ODIR-N;\@ $n_{\text{val}} = 470, n_{\text{test}} = 458$.}
      \label{tab:odir-n-confidence-intervals}
    \end{subtable}
    
    \vphantom{M}
    
    \begin{subtable}[h]{\linewidth}
      \centering
      \begin{tabular}{lrrrccrrrr}
      \toprule
      run & epoch & \multicolumn{2}{c}{$\AUC$} & \multicolumn{2}{c}{$\mathrm{CI}_{\alpha}$} & \multicolumn{2}{c}{$p_{\text{empir}}$} & \multicolumn{2}{c}{$p_{\text{adj}}$} \\
          &       &      val & test & val & test & val & test & val & test \\
      \midrule
      C1 & $16$ & $0.63^*$ &  $0.60^*$ & $(0.58, 0.69)$ & $(0.54, 0.65)$ & $<0.001$ &  $<0.001$ & $<0.0011$ & $<0.0011$\\
      C2 &  $4$ & $0.63^*$ &  $0.60^*$ & $(0.58, 0.69)$ & $(0.54, 0.66)$ & $<0.001$ &  $<0.001$ & $<0.0011$ & $<0.0011$\\
      C3 &  $5$ & $0.63^*$ &  $0.58^*$ & $(0.56, 0.68)$ & $(0.52, 0.63)$ & $<0.001$ &  $0.007$  & $<0.0011$ & $0.0071$\\
      C4 & $30$ & $0.63^*$ &  $0.60^*$ & $(0.57, 0.69)$ & $(0.55, 0.66)$ & $<0.001$ &  $0.001$  & $<0.0011$ & $0.0011$\\
      C5 &  $5$ & $0.62^*$ &  $0.55^\texttt{\#}$ & $(0.56, 0.67)$ & $(0.49, 0.61)$ & $<0.001$ &  $0.055$  & $<0.0011$ & $0.0550$\\
      C6 &  $5$ & $0.65^*$ &  $0.58^*$ & $(0.59, 0.70)$ & $(0.52, 0.64)$ & $<0.001$ &  $0.006$  & $<0.0011$ & $0.0062$\\
      \bottomrule
      \end{tabular}
      \caption{\centering ODIR-C;\@ $n_{\text{val}} = 381, n_{\text{test}} = 380$.}
      \label{tab:odir-c-confidence-intervals}
    \end{subtable}

    \caption{Single-domain results for fine-tuned ResNet-152 models.
      $B = 1000, \alpha = 0.05$. Unless explicitly stated,
      $p_{\text{empir}} \leq 10^{-3}$ for all significance tests prior to
      correction for multiple comparisons, and
      $p_{\text{adj}} \leq 1.1 \cdot 10^{-3}$ for all significance tests after
      adjustment for multiple comparisons using {BH}. Significant results are marked by an asterisk. Trend for significance ($0.05<p<0.1$) is marked by a number sign.}

       \label{tab:confidence-intervals}
  \end{adjustwidth}
  
\end{table}

\subsection{Ensembling}
\label{sec:results-naive-ensembling}

In this section, we describe results corresponding to the $(10, 20)$-ensemble
classifier described in~\nameref{sec:ensembling-method}. In particular, we
include in this section the results pertaining to the validation and test AUC
performance of $10$ models developed using the training and validation
partitions of the DOVS-ii database, E1--10. The models E1--10 attained a median
test AUC score of $0.69$ and scores ranged between $0.67$ and $0.71$. The median
validation AUC score was $0.75$ and scores ranged between $0.74$ and $0.76$,
suggesting negligible extent of overfit to the validation partition. All but one
model run converged within the first $49$ epochs, as displayed in the ``epoch''
column of~\autoref{tab:conf-stats-ensemble}. The ensemble classifier attained an
AUC score of $0.72$ on the test partition of the DOVS-ii database, higher than
any test score for the constituent models. Its AUC score on the validation
partition was $0.79$, suggesting some level of overfit to the validation set
during model development.

The adjusted $p$-values $p_{\text{adj}}$ were computed using $B = 1000$
bootstrap replicates with sample sizes $n_{\text{val}} = 214$ and
$n_{\text{test}} = 212$. For each of the tests for significance,
$p_{\text{adj}}$ was found to satisfy $p_{\text{adj}} \leq 1.1 \cdot 10^{-3}$,
implying the results were found to be significant at an $\alpha = 0.05$
confidence level. Confidence intervals for the test AUC score for the $11$
models were computed at the $\alpha = 0.05$ confidence level. Amongst the
component models, the confidence intervals were similar. The confidence interval
for the ensemble test AUC score was found to be $(0.67, 0.77)$. All values are shown
in~\autoref{tab:dovs-i-confidence-intervals}.

\begin{table}
  \centering
  \begin{tabular}{rrrrcc}
    \toprule
    run & epoch & \multicolumn{2}{c}{$\AUC$} & \multicolumn{2}{c}{$\mathrm{CI}_{\alpha}$} \\
        &       &      val & test & val & test \\
    \midrule
    E1 & $45$ & $0.76^*$ & $0.71^*$ & $(0.71, 0.81)$ & $(0.66, 0.76)$ \\ 
    E2 & $34$ & $0.76^*$ & $0.69^*$ & $(0.71, 0.81)$ & $(0.63, 0.74)$ \\ 
    E3 & $63$ & $0.75^*$ & $0.71^*$ & $(0.70, 0.80)$ & $(0.66, 0.77)$ \\ 
    E4 & $13$ & $0.75^*$ & $0.70^*$ & $(0.70, 0.79)$ & $(0.65, 0.75)$ \\ 
    E5 & $6$ & $0.75^*$ & $0.67^*$ & $(0.70, 0.79)$ & $(0.61, 0.72)$ \\ 
    E6 & $49$ & $0.75^*$ & $0.70^*$ & $(0.70, 0.80)$ & $(0.65, 0.75)$ \\ 
    E7 & $10$ & $0.74^*$ & $0.69^*$ & $(0.69, 0.79)$ & $(0.63, 0.74)$ \\ 
    E8 & $17$ & $0.74^*$ & $0.69^*$ & $(0.69, 0.79)$ & $(0.64, 0.74)$ \\ 
    E9 & $21$ & $0.74^*$ & $0.68^*$ & $(0.70, 0.79)$ & $(0.63, 0.74)$ \\ 
    E10 & $11$ & $0.74^*$ & $0.69^*$ & $(0.69, 0.79)$ & $(0.63, 0.74)$ \\ 
    E$^{*}$ & --- & $0.79^*$ & $0.72^*$ & $(0.74, 0.84)$ & $(0.67, 0.77)$ \\

    \bottomrule
  \end{tabular}
  \caption{Ensembling experiment results: scores for E1--10 developed
    using the DOVS-ii database, and the ensemble classifier $E^{*}$. The
    unadjusted $p$-value obtained from bootstrap replicates satisfies
    $p_{\text{empir}} \leq 10^{-3}$ for each line in the table. The adjusted
    $p$-value $p_{\text{adj}}$ satisfies $p_{\text{adj}} \leq 1.1 \times 10^{-3}$
    ($B=1000, \alpha=0.05, n_{\text{test}} = 376$). The row for run E$^{*}$
    displays statistics for the $(10, 20)$-ensemble classifier created from the
    component models E1--10. Significant results are marked by an asterisk.}
  \label{tab:conf-stats-ensemble}
\end{table}

\subsection{Domain adaptation results}
\label{sec:results-domain-adaptation}

In this section we describe results pertaining to the domain adaptation
experiments, including scores, confidence intervals and significance values. All
values pertaining to these experiments are given
in~\autoref{tab:cross-test-confidence-stats}. Specifically, for the AUC scores
and associated significance scores of D1--5 evaluated on the ODIR-N and ODIR-C
datasets, see~\autoref{tab:cross-test-Di}; for those pertaining to N1--6
evaluated on the DOVS-ii dataset, see~\autoref{tab:cross-test-Ni}; and for
those pertaining to C1--6 evaluated on the DOVS-ii dataset,
see~\autoref{tab:cross-test-Ci}.

When tested on the ODIR-N and ODIR-C datasets, models D1--5 attained a median
AUC score of $0.564$. AUC scores ranged between $0.555$ and $0.572$ and were all
significantly greater than chance-level performance
($p_{\text{adj}} \leq 1.1\cdot 10^{-3}$ for $B = 1000$ bootstrap replicates with
$n_{\text{ODIR-N}} = 3098$ and $n_{\text{ODIR-C}} = 2577$). Confidence intervals
for these results were computed at an $\alpha = 0.05$ confidence level and
appear in~\autoref{tab:cross-test-Di}. The difference in scores between tests on
ODIR-N and ODIR-C are minimal; for each model the confidence intervals
associated to the two scores typically have a high degree of overlap.

When tested on the DOVS-ii dataset, models N1--6 attained a median AUC score of
$0.586$.  AUC scores ranged between $0.580$ and $0.607$ and all were
significantly greater than chance ($p_{\text{adj}} \leq 1.1\cdot 10^{-3}$ for
$B = 1000$ bootstrap replicates with $n_{\text{ODIR-N}} = 2496$). Confidence
intervals for these results were computed at an $\alpha = 0.05$ confidence level
and appear in~\autoref{tab:cross-test-Di}. When tested on the DOVS-ii database,
models C1--6 attained a median AUC score of $0.61$. AUC scores ranged between
$0.574$ and $0.617$ and all were significantly greater than chance
($p_{\text{adj}} \leq 1.1\cdot 10^{-3}$ for $B = 1000$ bootstrap replicates with
$n_{\text{ODIR-N}} = 2496$). Confidence intervals for these results were
computed at an $\alpha = 0.05$ confidence level and appear
in~\autoref{tab:cross-test-Di}. Despite being developed using a smaller dataset,
C1--6 overall achieved greater test AUC compared to N1--6, though their
performance is similar.


\begin{table}[h!]
  \begin{adjustwidth}{-1.25in}{.6in}
    \centering
    \begin{subtable}[h]{\textwidth}
      \centering
      \begin{tabular}{lrrrcc}
        \toprule
      run &  epoch &  \multicolumn{2}{c}{$\AUC$} & \multicolumn{2}{c}{$\mathrm{CI}_{\alpha}$} \\
          &        & ODIR-N & ODIR-C & ODIR-N & ODIR-C  \\
      \midrule
      D1 &    $107$ &   $0.555^*$ & $0.563^*$ &  $(0.535, 0.577)$ & $(0.540, 0.585)$ \\
      D2 &     $37$ &   $0.565^*$ & $0.572^*$ &  $(0.544, 0.585)$ & $(0.551, 0.594)$ \\
      D3 &     $46$ &   $0.569^*$ & $0.572^*$ &  $(0.550, 0.589)$ & $(0.550, 0.594)$ \\
      D4 &     $60$ &   $0.558^*$ & $0.555^*$ &  $(0.537, 0.580)$ & $(0.533, 0.577)$ \\
      D5 &     $73$ &   $0.562^*$ & $0.566^*$ &  $(0.541, 0.581)$ & $(0.543, 0.588)$ \\
        \bottomrule
      \end{tabular}
      \caption{\centering DOVS-i models;
        $n_{\text{ODIR-N}} = 3098, n_{\text{ODIR-C}} =
        2577$.\label{tab:cross-test-Di}}
    \end{subtable}

    \vphantom{M}
    
    \begin{subtable}[h]{\textwidth}
      \centering
      \begin{tabular}{lrrrcc}
        \toprule
        run & epoch & \multicolumn{2}{c}{$\mathrm{AUROC}$} & \multicolumn{2}{c}{$\mathrm{CI}_{\alpha}$} \\
            &        &     ODIR-N     &     ODIR-C         &      ODIR-N     &        ODIR-C  \\

        \midrule
        E1      & $45$ &  $0.567^{*}$ & $0.580^{*}$ & $(0.549, 0.589)$ & $(0.560, 0.604)$  \\
        E2      & $34$ &  $0.578^{*}$ & $0.590^{*}$ & $(0.560, 0.598)$ & $(0.569, 0.612)$  \\
        E3      & $63$ &  $0.559^{*}$ & $0.569^{*}$ & $(0.540, 0.579)$ & $(0.546, 0.591)$  \\
        E4      & $13$ &  $0.535^{*}$ & $0.542^{*}$ & $(0.514, 0.554)$ & $(0.521, 0.564)$  \\
        E5      &  $6$ &  $0.571^{*}$ & $0.569^{*}$ & $(0.550, 0.591)$ & $(0.545, 0.590)$  \\
        E6      & $49$ &  $0.553^{*}$ & $0.575^{*}$ & $(0.532, 0.572)$ & $(0.552, 0.596)$  \\
        E7      & $10$ &  $0.558^{*}$ & $0.557^{*}$ & $(0.538, 0.578)$ & $(0.535, 0.580)$  \\
        E8      & $17$ &  $0.549^{*}$ & $0.569^{*}$ & $(0.529, 0.569)$ & $(0.548, 0.589)$  \\
        E9      & $21$ &  $0.540^{*}$ & $0.553^{*}$ & $(0.521, 0.559)$ & $(0.531, 0.575)$  \\
        E10     & $11$ &  $0.537^{*}$ & $0.556^{*}$ & $(0.518, 0.557)$ & $(0.535, 0.579)$ \\
        E$^{*}$ &      &  $0.562^{*}$ & $0.580^{*}$ & $(0.542, 0.584)$ & $(0.558, 0.603)$ \\
        \bottomrule
      \end{tabular}
      \caption{\centering DOVS-ii models;
        $n_{\text{ODIR-N}} = 3098, n_{\text{ODIR-C}} = 2577$. For E4,
        $p_{\text{empir}} = 10^{-3}$.\label{tab:cross-test-Dii}}
    \end{subtable}

    \vphantom{M}
    
    \begin{subtable}[t]{.45\textwidth}
      \centering
      \begin{tabular}{lrrc}
        \toprule
      run &  epoch &  $\AUC$ & $\mathrm{CI}_{\alpha}$ \\
          &        &  DOVS-ii          &   DOVS-ii \\
      \midrule
      N1 &     $43$ &  $0.580^*$ &  $(0.558, 0.604)$ \\
      N2 &     $47$ &  $0.584^*$ &  $(0.563, 0.606)$ \\
      N3 &     $36$ &  $0.607^*$ &  $(0.585, 0.628)$ \\
      N4 &     $36$ &  $0.588^*$ &  $(0.566, 0.610)$ \\
      N5 &     $21$ &  $0.600^*$ &  $(0.579, 0.621)$ \\
      N6 &     $66$ &  $0.583^*$ &  $(0.561, 0.606)$ \\
        \bottomrule
      \end{tabular}
      \caption{\centering ODIR-N models; $n_{\text{DOVS-ii}} = 2496$. \label{tab:cross-test-Ni}}
    \end{subtable}
    \qquad\qquad
    \begin{subtable}[t]{.45\textwidth}
      \centering
      \begin{tabular}{lrrc}
        \toprule
      run &  epoch & $\AUC$ & $\mathrm{CI}_{\alpha}$ \\
          &        &  DOVS-ii          &   DOVS-ii \\
      \midrule
      C1 &     $16$ &  $0.606^*$ &  $(0.586, 0.630)$ \\
      C2 &      $4$ &  $0.574^*$ &  $(0.553, 0.595)$ \\
      C3 &      $5$ &  $0.616^*$ &  $(0.593, 0.636)$ \\
      C4 &     $30$ &  $0.617^*$ &  $(0.595, 0.638)$ \\
      C5 &      $5$ &  $0.579^*$ &  $(0.557, 0.601)$ \\
      C6 &      $5$ &  $0.606^*$ &  $(0.584, 0.627)$ \\
        \bottomrule
      \end{tabular}
      \caption{\centering ODIR-C models; $n_{\text{DOVS-ii}} = 2496$. \label{tab:cross-test-Ci}}
    \end{subtable}
    
    \caption{Domain adaptation results.  $B = 1000$,
      $\alpha = 0.05$. Unless explicitly stated, the unadjusted $p$-values
      satisfy $p_{\text{empir}} \leq 10^{-3}$ for all significance tests prior
      to correction for multiple comparisons, and
      $p_{\text{adj}} \leq 1.1 \cdot 10^{-3}$ for all significance tests after
      adjustment for multiple comparisons using {BH}. Significant results are marked by an asterisk.}
    \label{tab:cross-test-confidence-stats}
  \end{adjustwidth}
\end{table}

\subsection{Model comparison}
\label{sec:results-model-comparison}

\begin{figure}[h]
  \centering
  \includegraphics[width=.8\textwidth]{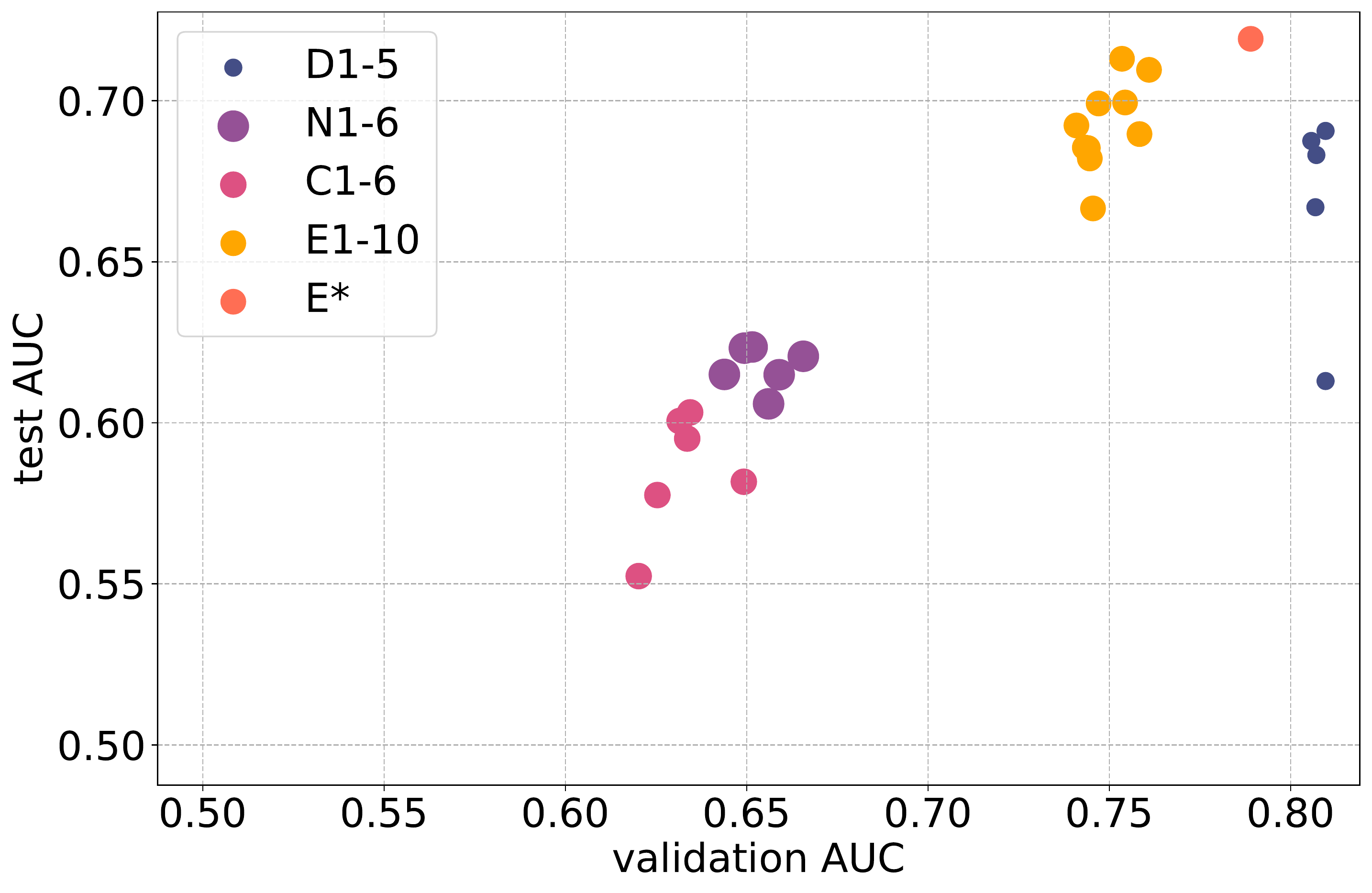}
  \caption{A graphical representation of all models developed in this work. On
    the $x$-axis is a model's AUC score on the validation partition of the
    relevant database; on the $y$-axis, its AUC score on the test
    partition. Points labelled D1--5 correspond with models trained and
    evaluated on DOVS-i; E1--10, DOVS-ii; N1--6, ODIR-N;\@ C1--6, ODIR-C. The
    ensemble model is denoted E$^{*}$. Marker size
    corresponds with the number of images in the database used for model
    development (\emph{cf.}~\autoref{tab:dataset-statistics}).}
  \label{fig:val-test-AUC-all-models}
\end{figure}

\begin{figure}[h]
  \centering
  \includegraphics[width=.8\textwidth]{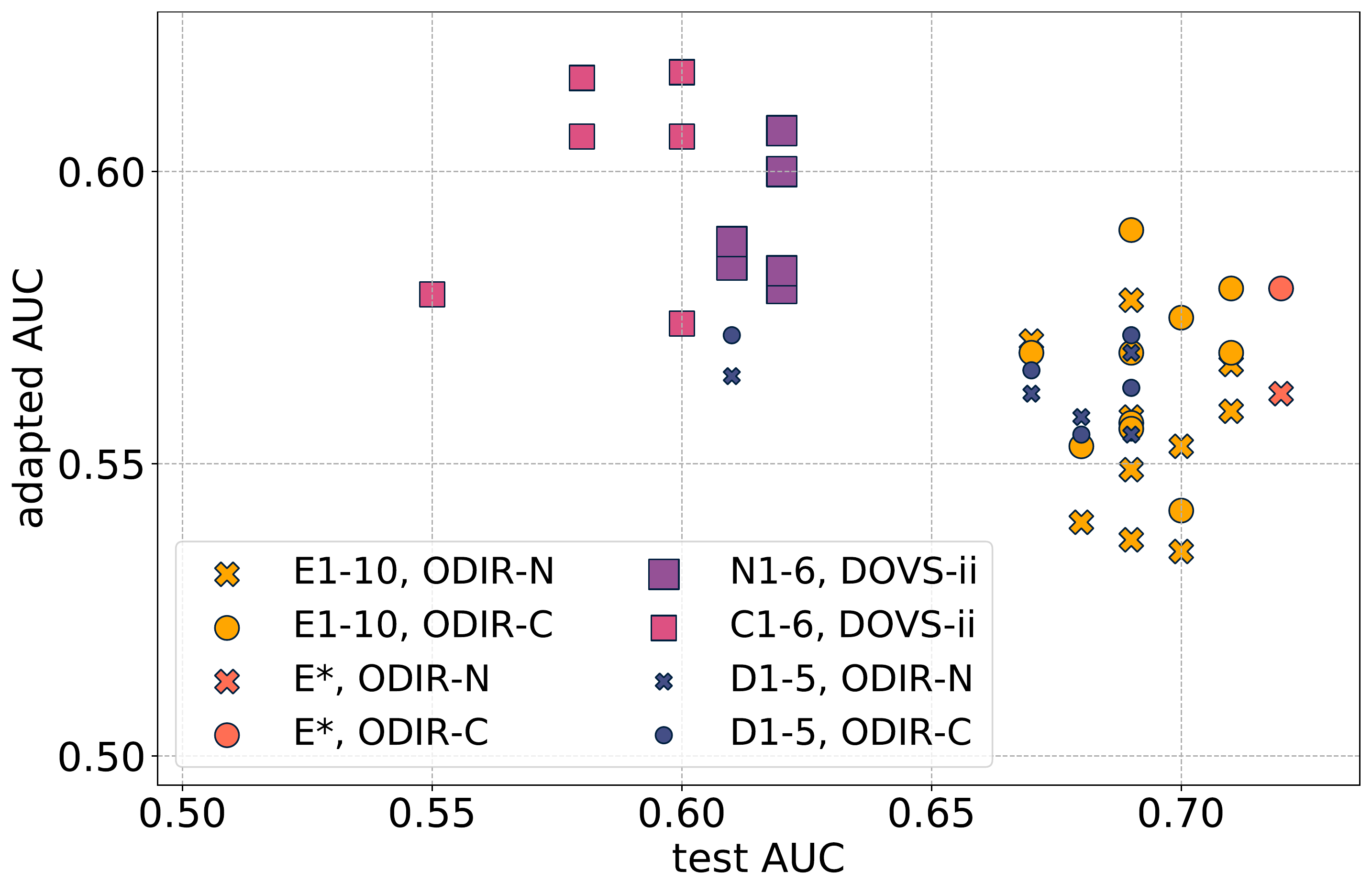}
  \caption[Graphical domain adaption results.]{A graphical representation of the
    domain adaptation results for the models developed in this work. On the
    $x$-axis is a model's AUC score on the test partition from the database on
    which it was trained; on the $y$-axis, its domain-adapted AUC score. Points
    labelled D1--5 correspond with models trained and evaluated on DOVS-i;
    E1--10, DOVS-ii; N1--6, ODIR-N;\ C1--6, ODIR-C. The ensemble model is denoted
    E$^{*}$. The marker radius of the plotted points corresponds with the number
    of images in the database used for model development
    (\emph{cf.}~\autoref{tab:dataset-statistics}). Each point is coloured
    according to the database on which the associated model was trained, and the
    shape of each point corresponds with the database on which that model was
    evaluated for its domain-adapted AUC score. }
  \label{fig:crosstest-auc-all-models}
\end{figure}

A summary overview of all models developed in the study is shown
in~\autoref{fig:val-test-AUC-all-models}
and~\autoref{fig:crosstest-auc-all-models}. In~\autoref{fig:val-test-AUC-all-models}
validation $\AUC$ is plotted against test $\AUC$. For example, models D1--5 were
developed using the training and validation partitions of the DOVS-i dataset;
their AUC score on the test partition is plotted as a function of their AUC
score on the validation partition. Only the retained subset (i.e., top 10) of
E1--20 are depicted (labeled E1--10), as the other models were discarded.
during development of the ensemble (\emph{cf.}~\nameref{sec:ensembling-method}).

This plot enables a visual comparison of model performance on different
datasets. In particular, it enables a direct comparison of models' test AUC
scores, the variability of test AUC, and the extent of overfitting. For example,
C1--6 generally attained the lowest validation and test scores, with N1--6 only
slightly better. In particular, models performed better on the single domain
task when developed using the DOVS datasets.

The ensemble E$^{*}$ attained the highest test score. Importantly, its
validation score is less than that for D1--5, meaning that the extent of
overfitting for E$^{*}$ is less. Its extent of overfitting is comparable with
its constituent models E1--10. Indeed, except for models D1--5, we interpret
that all models overfit by approximately $0.05$. Finally, we observe that E1--10
have higher test AUC and overfit less than D1--5. In particular, the models
developed with the larger DOVS-ii dataset performed better than those developed
using DOVS-i.

For those models for which there were multiple runs, we observe that the
variability of the test AUC was comparable. Namely, the lowest and highest test
AUC scores for a group of models was generally within $0.05$, except one outlier
each for D1--5 and C1--6.

In~\autoref{fig:crosstest-auc-all-models}, test $\AUC$ is plotted against
adapted $\AUC$, meaning the $\AUC$ score that was achieved on the domain
adaptation task for the given (model, dataset) pair. Two clusters are
discernible from the plot: the left cluster is comprised mainly of N1--6 and
C1--6 models tested on DOVS-ii, who achieve lower test $\AUC$ scores (compared
to the other discernible cluster), but adapt better to out-of-distribution data
(\ie DOVS-ii). The other discernible cluster is comprised of models trained on
DOVS-i or DOVS-ii. We observe that the ensemble model E$^{*}$ acheives the
highest test $\AUC score$ and one of the highest adapted $\AUC$ scores on both
ODIR-N and ODIR-C.



\section{Discussion}
\label{sec:discussion}

\subsection{Training metrics}
\label{sec:discussion-training-metrics}

Despite the inherently stochastic nature of the training process, it can be seen
(\autoref{fig:dovs-i-training-metrics}) that each model run's performance is
comparable to another's. That a model achieves similar performance from
realization to realization provides intuition that the training algorithm is
behaving appropriately.

It is curious that the validation BCE score underwent an overall \emph{increase}
for all three databases. We observed this phenomenon consistently throughout our
experiments. We believe it to be the result of complex interplay between the data
distribution of retinal images, the complexity of the classification task, and
the choice of loss function. 

Our supplementary experiment (\nameref{S1_loss}) compared two loss functions in
an effort to shed light on this phenomenon: (1) binary cross entropy loss, and
(2) an alternate loss function that is balanced in the way penalties change on
the two halves of the curve straddling the threshold of 0.5. The results of this
supplementary experiment indicate that both models show a tendency to generate a
small number of incorrect but confident responses as training
progresses. However, the phenomenon of increasing loss was no longer apparent
when cross-entropy loss was replaced with the ``balanced'' loss function (see
\nameref{S1_loss}). A rigorous causal understanding of the behaviour that
eventually yields a model producing outlier confident incorrect classifications
for both categories is beyond the scope of the present work, and would make for
an involved subject of further study.

\subsection{Model-database duality}
\label{sec:discussion-model-db-comparison}

We observe that models perform better on the DOVS database in single-domain
tasks. In particular, D1--5 achieve higher validation metrics during training
(\emph{cf.}~\autoref{fig:dovs-i-training-metrics}) and have greater test AUC
(\emph{cf.}~\autoref{tab:confidence-intervals}). Furthermore, N1--6 and C1--6
attained greater AUC when evaluated on the DOVS-ii dataset, relative to D1--5 on
either ODIR-N or ODIR-C (\emph{cf.}~\autoref{tab:confidence-intervals}). In
fact, such results may initially seem surprising: shouldn't the higher-scoring
models be more likely to perform well when adapted to new datasets? In fact, we
believe this behaviour underscores how underlying structure in the data is
critical to model performance. The ODIR database~\cite{odir-database} was
obtained from many different hospitals and medical centres using various cameras
(produced by different companies), resolutions and image qualities. The effects
of this variation are two-fold. First, this variation increases the complexity
of the ODIR database, likely increasing the difficulty of the learning
task. This foreshadows lower scores like those seen in
\nameref{sec:results-prototypical}. On the other hand, when using a sufficiently
complex model, training on a dataset of greater complexity has the opportunity
to produce a more robust model. This point of view is consistent with N1--6 and
C1--6 achieving higher scores on the DOVS-ii database than D1--5 did on ODIR-N
or ODIR-C. This is also consistent with the observation that D1--5 typically
achieve higher scores when evaluated on the ODIR-C dataset than on the ODIR-N
dataset.

In~\nameref{sec:results-domain-adaptation}, we observed that models N1--6 and
C1--6 attained test AUC scores above $0.6$
(see~\autoref{tab:cross-test-Ni},~\ref{tab:cross-test-Ci}, respectively), while
none of the models D1--5 attained this (arbitrarily chosen) threshold
(\autoref{tab:cross-test-Di}), nor did E1--10, E$^{*}$
(\autoref{tab:cross-test-Dii}). We offer the following \emph{ex post facto}
justification for this observation, which supports the description provided in
the previous paragraph. Namely, the models fine-tuned on a relatively more
complex dataset were able to generalize better to unfamiliar data than were the
models developed using a more uniform database (DOVS).


Importantly, in~\autoref{fig:crosstest-auc-all-models}, we observed two
interesting facts about ODIR-N \emph{vs.} ODIR-C that are suggestive about the
role of image quality in dataset curation and model development. First, compared
to ODIR-N, models trained on ODIR-C generally achieved higher adapted $\AUC$
scores, even though their test $\AUC$ scores were uniformly lower. Second,
D1--5, E1--10 and E$^{*}$ achieved better adapted $\AUC$ scores when tested on
ODIR-C than when tested on ODIR-N. Together, these two observations suggest the
importance of clean data when curating datasets for model development; and for
ensuring efficacy in domain adaptation tasks (such as deploying a developed
model in a new clinical setting).

\subsection{Ensembling}
\label{sec:discussion-ensembling}

In this work, we investigated an ensemble generated by averaging model
beliefs. We found that model averaging improved classifier performance, with
only moderate increase in the computational complexity of the problem. Indeed,
typically the development of a deep learning model involves several rounds of
training to allow for hyperparameter tuning and changes to the model
architecture. Thus, adequately performing models found during this development
process are apt candidates to comprise an ensemble classifier.

Several interesting avenues present themselves for further study. For instance,
we do not investigate how the ensemble could be tuned to produce improved
results. For example, the ensemble could have been computed as a weighted sum of
the individual networks, whose weights were determined by the validation AUC
of each model; or through a voting procedure. Alternatively, the tuning
procedure could have been learned during a second round of training. Such
approaches are typically more suitable when larger databases are
available. Finally, we do not examine dependence of AUC on $\ell$ or $L$. This
subject would be an involved investigation that would be interesting to explore
in future work.


\section*{Conclusion}

In this work we have exhibited how deep learning models can achieve super-human
performance on challenging retinal image analysis tasks even when only small
databases are available. Two previous studies have reported successful
classification of sex from retinal fundus images using deep learning, though
this was achieved via use of much larger datasets in each case: over $80\,000$
images in~\citet{korot2021predicting} and over 1.5 million images
in~\citet{poplin2018prediction}. Further, we have demonstrated that deep learning
models for retinal image analysis can succeed at transductive transfer learning
when applied to a similar task on a different data distribution. Finally, we
have exhibited how the model development process lends itself to the creation of
ensemble classifiers composed of deep learning models, whose performance can
exceed that of its constituent models.

Data presents as the primary bottleneck in machine learning applications of medical image analysis, with reductions in dataset size often associated with reductions in model performance. In this proof-of-concept study, we have shown that small datasets can be leveraged to obtain meaningful performance in the context of retinal fundus image classification. Our study has revealed several important considerations when data set size is a limitation, such as use of transfer learning in model development, and use of ensembling to maximize classification performance. These results have also highlighted maintaining strict image quality criteria as a strategy to improve generalization in applications that involve domain adaptation. These findings move us one step further in the democratization and applicability of deep learning in medical imaging.



\section*{Acknowledgments}



\bibliography{dlri}

\clearpage
\appendix

\section*{S1 Methods}
\label{S1_methods}

\subsection*{Neural networks}
\label{S1_neural_networks}

A deep neural network is a highly parametrized function that is useful for
general purpose function approximation. Suppose an unknown ``ground truth''
function $f : \reals^{d} \to \{0, 1\}$ describes a labeling scheme for points
$x \in \reals^{d}$ (\ie $y = F(x)$ is the ``label'' for the point $x$). Further
suppose one obtains several labeled realizations of data from this function:
$\mathcal{S} := \{(x_{1}, y_{1}), \ldots, (x_{n}, y_{n})\}$, where
$y_{i} = F(x_{i})$, $i \in [n]$. It is often desirable to be able to
approximate the function $F$ from the data $\mathcal{S}$, and in several
complex, modern scenarios, the state-of-the-art choice is to train a deep
neural network.

Let $\mathcal{H}$ denote a hypothesis class. For example, $\mathcal{H}$ may
describe a particular deep neural network architecture (or set of
architectures). The goal of training a deep neural network is, given
$\mathcal{S} \subseteq \reals^{d}$, to obtain
$h \in \mathcal{H}, h : \reals^{d} \to [0, 1]$ that approximates well the
action of $F$ on $x_{i}, i \in [n]$. In particular, given a convex function
$\ell : [0, 1] \times \{0, 1\} \to \reals_{+}$, find $h \in H$ so that
$\ell(h(x), y) \ll 1$ for all $(x, y) \in \mathcal{S}$.

\subsection*{Metrics}
\label{S1_metrics}

\subsubsection*{Binary cross-entropy}
\label{S1_bce}

Binary cross-entropy was the metric used to optimize the model --- which is to
say that the ``misfit'' that was approximately minimized during model training,
between model predictions and the true labels associated to each image, was 
binary cross-entropy.

\subsubsection*{Receiver Operating Characteristic}
\label{S1_roc}

Suppose a ``ground truth'' is given by the function
$f : \mathcal{X} \to \{-1, 1\}$. Further, suppose an experimenter develops a
hypothesis for the behaviour of $f$, in the form of a binary classifier
$h: \mathcal{X} \to \{-1, 1\}$. Given a finite subset
$\mathcal{S} \subseteq \mathcal{X}$, define the \textbf{true positive rate} and
\textbf{false positive rate} of the hypothesis on the dataset $\mathcal{S}$ by,
respectively,
\begin{align*}
  \mathrm{tpr}(h, \mathcal{S}) %
  &:= |P|^{-1} \sum_{x \in \mathcal{P}} \1(h(x) = 1)
  &
    \mathrm{fpr}(h, \mathcal{S}) %
  &:= |N|^{-1} \sum_{x \in \mathcal{N}} \1(h(x) = 1)
\end{align*}
Now assume that $h$ comes from a family of hypotheses,
$h_{\theta} : \mathcal{X} \times \Theta \to \{0, 1\}$. For example, perhaps
$h_{\theta}(\cdot) := \1\big(\tilde h(\cdot) > \theta)$ for some
$\tilde h : \mathcal{X} \to [0, 1]$, and $\Theta = [0, 1]$. In this case,
$\tilde h$ outputs a \emph{score} for the point $x$, and $\theta$ acts as a
\emph{threshold}. By a mild abuse of notation, we can define the true positive
rate and false positive rate as functions of the threshold $\theta$:
\begin{align*}
  \mathrm{tpr}(\theta) %
  &:= |P|^{-1} \sum_{x \in \mathcal{P}} \1(h_{\theta}(x) = 1)
  &
    \mathrm{fpr}(\theta) %
  &:= |N|^{-1} \sum_{x \in \mathcal{N}} \1(h_{\theta}(x) = 1)
\end{align*}
The \textbf{receiver operating characteristic} (ROC) of the function $h_{\theta}$ on the dataset $\mathcal{S}$ is given as the 
\begin{align*}
  \mathrm{ROC}(h; \mathcal{S}, \Theta) %
  &:= \{ (\mathrm{fpr}(\theta), \mathrm{tpr}(\theta)) : \theta \in \Theta\}
\end{align*}
We can view $\mathrm{ROC}(h; \mathcal{S}, \Theta)$ as a graph, with the true
positive rate changing as a function of the false positive rate. If $h$ is a
good model for $f$, then the true positive rate and true negative rate are
high. Now, $\mathrm{fpr} = 1 - \mathrm{tnr}$, where $\mathrm{tnr}$ is the true
negative rate. In particular, the true positive rate is high and the false
positive rate is close to zero. Also observe that $\mathrm{tpr}$ is a
monotonically non-decreasing function of $\mathrm{fpr}$. Thus, if $h$ is a good
model for $f$, then the area under the curve of the ROC should be large. To measure this area, we simply integrate as follows:
\begin{align}
  \label{eq:def-auroc}
  \AUC (h; \mathcal{S}) := \int_{0}^{1} \operatorname{tpr}(\operatorname{fpr}^{-1}(\theta)) \d \theta.
\end{align}

\subsection*{Network architecture}
\label{S1_network-architecture}

The network architecture used in this work is modified from a deep residual
network, ResNet-152, originally developed in~\cite{he2016deep}. This network is
composed of \emph{residual units} $\mathcal{H}(x)$, which resemble the
following:
\begin{align*}
  \mathcal{H}(x) := x + \mathcal{F}(x), %
  \quad \text{where} \quad %
  \mathcal{F}(x) %
  := W_{2}\sigma(W_{1} x + b_{1}) + b_{2}. 
\end{align*}
One residual unit may be composed with a subsequent residual unit as
$\tilde{\mathcal{H}}(x) := \mathcal{H}^{2}(\sigma(\mathcal{H}^{1}(x)))$, where,
as above, $\sigma$ is an activation function. In the implementation, there are
typically more than two matrices $W_{i}$ in each unit; the matrices $W_{i}$
typically correspond to a set of convolutions of a fixed filter size (\eg $64$
convolution filters of size $3 \times 3$); and the activation function $\sigma$
is typically \texttt{ReLU}. Moreover, the network may not be entirely composed
of residual units: portions resembling more classical architectures may be used
as well. For example, it is common to append one or more fully connected layers
to a residual network: if $f_{1}, f_{2}$ are fully connected layers, so that
$f_{i}(x) = \tilde W_{i} x + \tilde b$ for $i \in [3]$ where $W_{i}$ and
$b_{i}$ are dense, then it is common to construct a network resembling
\begin{align*}
  \tilde{\mathcal{H}}(x) %
  := (f_{3} \circ f_{2} \circ f_{1} \circ \mathcal{H}^{k}\circ \sigma \circ \cdots %
  \circ \mathcal{H}^{2} \circ \sigma \circ \mathcal{H}^{1})(x). 
\end{align*}
For full implementation details of ResNet-152 we refer the reader
to~\cite{he2016deep} as well as the PyTorch~\cite{paszke2017automatic} source
code for the \texttt{ResNet} class, available in
\texttt{torchvision.models.resnet} available on
\href{https://github.com/pytorch/vision/blob/a91fe7221b55c55dbbc23c23aecf33d470a5c08e/torchvision/models/resnet.py#L119}{GitHub}.

The network architecture was modified from that of~\cite{he2016deep} by removing
the final layer of the network and appending $2$ fully connected layers, with
Dropout. The new layers of the network were randomly initialized according to
PyTorch \texttt{v1.01}~\cite{paszke2017automatic} default settings.


\section*{S1 Choice of loss}
\label{S1_loss}

\begin{figure}[h]
  \centering
  \includegraphics[width=\textwidth]{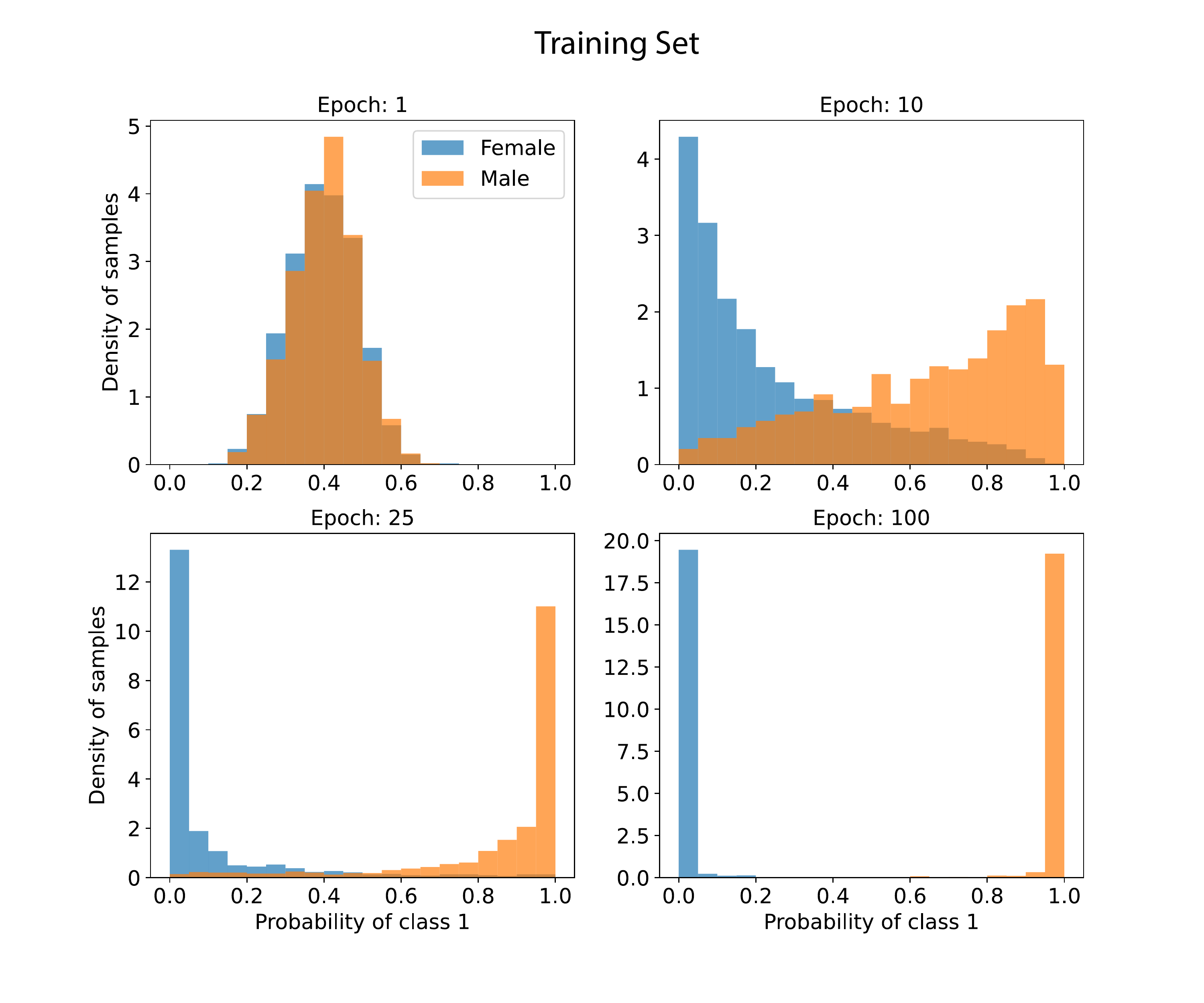}
  \caption{The histogram of predicted probabilities for training samples over training epochs: blue denotes female samples (class 0), orange denotes male samples (class 1).}
  \label{fig:odirN_training_histograms}
\end{figure}

While training-time binary cross-entropy (BCE) loss monotonically decreases on
the training partition, we observed that validation BCE loss increased for a
majority of the epochs, despite continued improvements on the accuracy and AUC
metrics (see Figures 1--3). To explain this behavior, we examined the output
probabilities of the model throughout the training epochs on ODIR-N. As shown
in~\autoref{fig:odirN_training_histograms}, training samples are not separable
during early epochs but become so: as training progresses the histograms
eventually have disjoint support. Ultimately, training samples are completely
separable and the model returns only probabilities close to zero or one, which
means it is \emph{confident} in its classifications on the training set.

\begin{figure}[h]
  \centering
  \includegraphics[width=\textwidth]{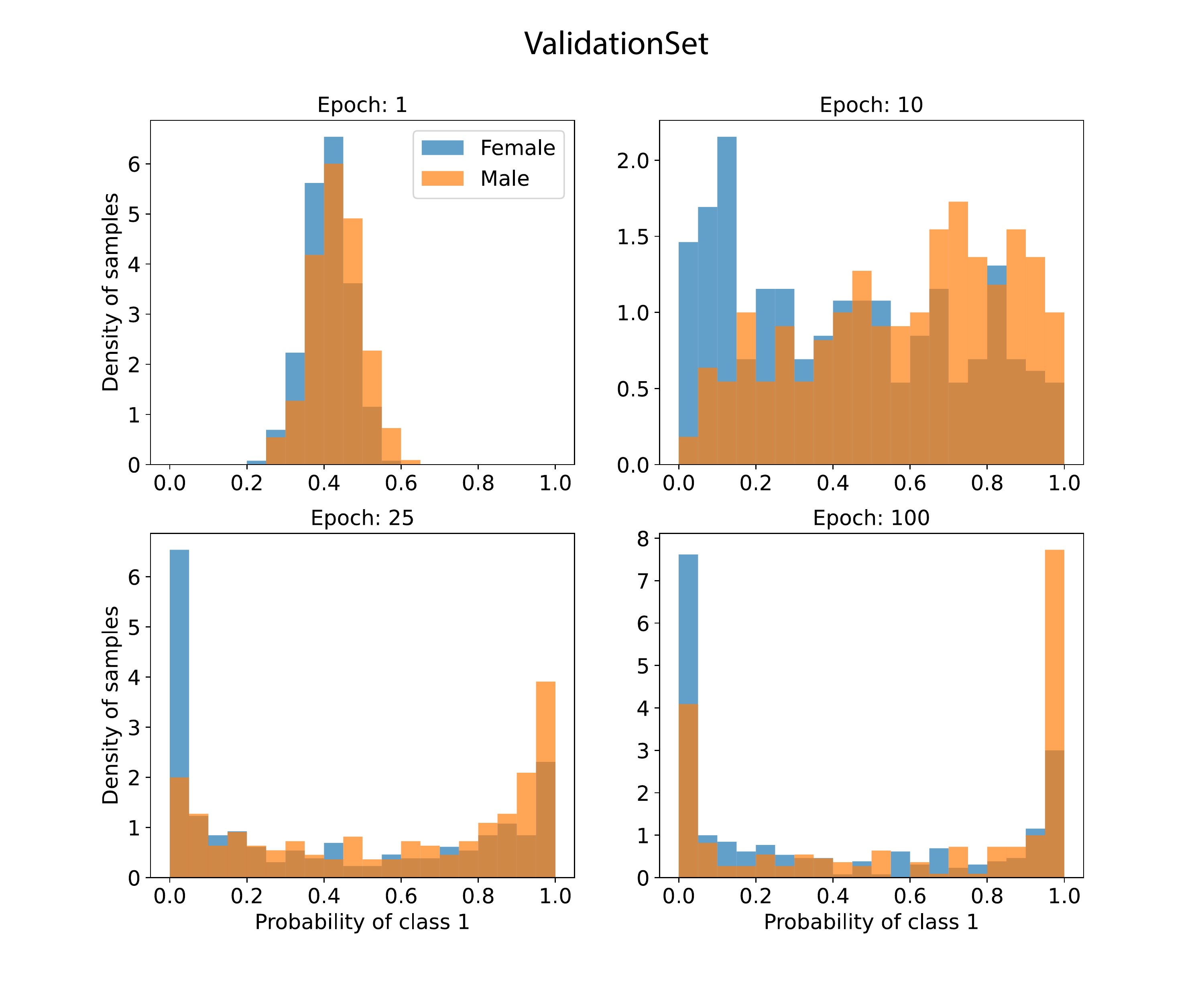}
  \caption{Histograms of predicted probabilities for validation samples,
    stratified by training epoch. Blue denotes female samples (class $0$),
    orange male (class $1$).}
  \label{fig:odirN_validation_histograms}
\end{figure}

Although the model's performance is not as high and the two histograms overlap,
the model insists on confident decisions on the validation partition
(see~\autoref{fig:odirN_validation_histograms}). Indeed, at the end of the
training process, the majority of the predicted probabilities are close to zero
or one. However, in contrast to the training set, some of the probabilities are
predicted incorrectly. By design, confident incorrect classifications incur
exponentially larger BCE loss. We argue, thereby, that the ``disproportionate''
loss incurred by confident incorrect classifications explains the training-time
behavior of the BCE loss on the validation partition.

\begin{figure}[h]
  \centering
  \captionsetup[subfigure]{justification=centering}
  \begin{subfigure}[h]{0.45\linewidth}
    \includegraphics[width=\textwidth]{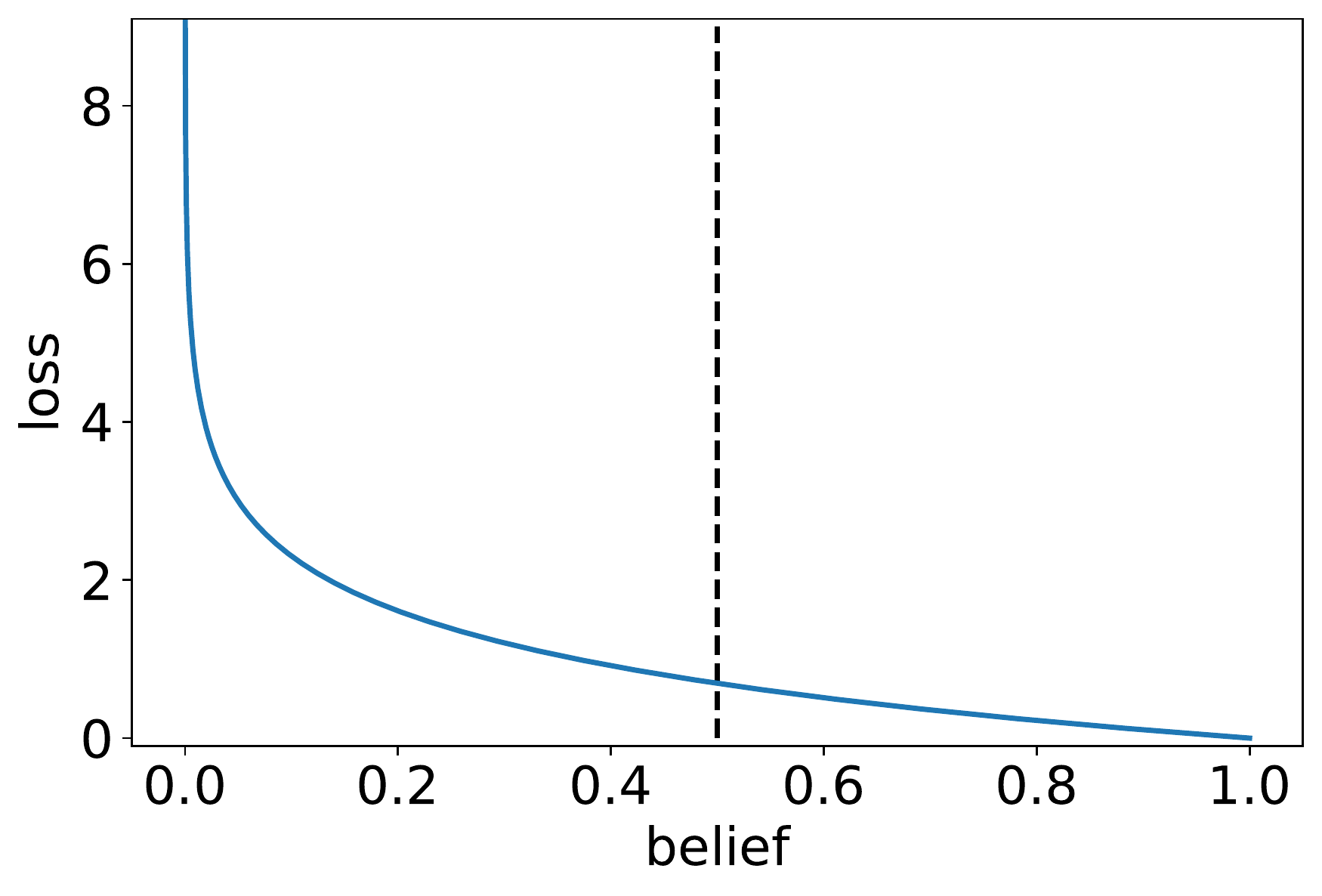}
    \caption{binary cross-entropy loss\label{fig:conditional-bce-loss}}
  \end{subfigure}\hfill
  \begin{subfigure}[h]{0.45\linewidth}
    \includegraphics[width=\textwidth]{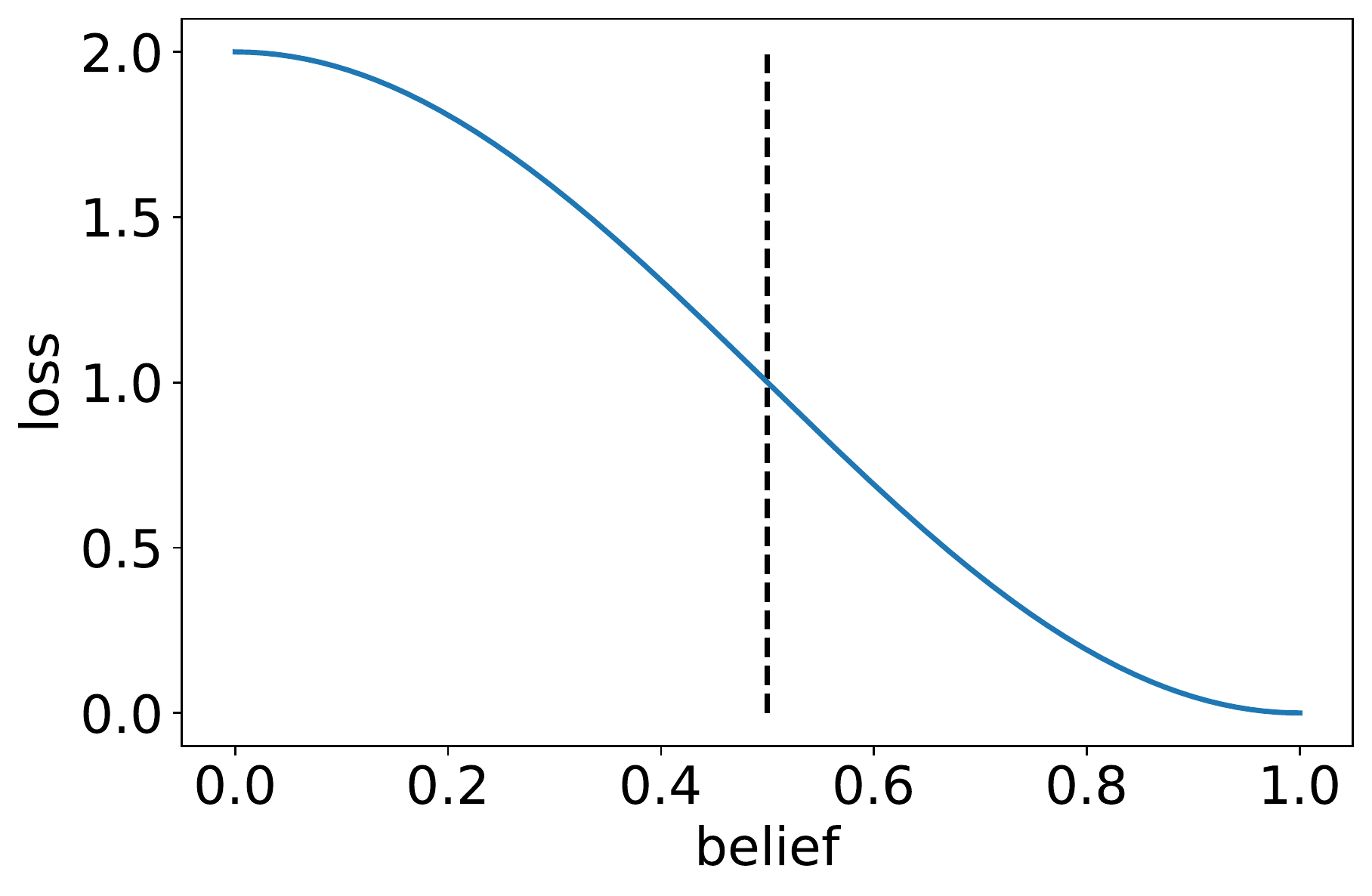}
    \caption{balanced loss\label{fig:conditional-balanced-loss}}
  \end{subfigure}
  \vspace{2pt}
  \caption{For each loss function, loss \emph{vs.} belief for class 1 when the
    true label is 1.}
  \label{fig:conditional-loss}
\end{figure}

In \autoref{fig:conditional-bce-loss}, BCE loss is plotted as a function of
predicted probability when the true label is one (\ie the model's \emph{belief}
that the true label is one). When the true label is predicted by the model with
probability one, the loss is zero, and as the probability decreases, the loss
increases exponentially. The model’s decision boundary is at 0.5 probability, at
which point, the predicted label switches. We can see that loss for
probabilities higher than this threshold (where the model predicts the true
label) is low and negligible. However, by definition of BCE, a much higher
penalty is incurred for incorrectly predicted probabilities, especially
probabilities around zero at which the loss function approaches
infinity. Therefore, although the two histograms
in~\autoref{fig:odirN_validation_histograms} separate during the training
epochs, the incorrect but confident predictions contribute large loss. This
increase in loss is not compensated by additional correct predictions in
progressive epochs; thus, we observe that loss increases for the validation set
during training.

\begin{figure}[h]
  \centering
  \captionsetup[subfigure]{justification=centering}
  \begin{subfigure}[h]{0.45\linewidth}
    \includegraphics[width=\textwidth]{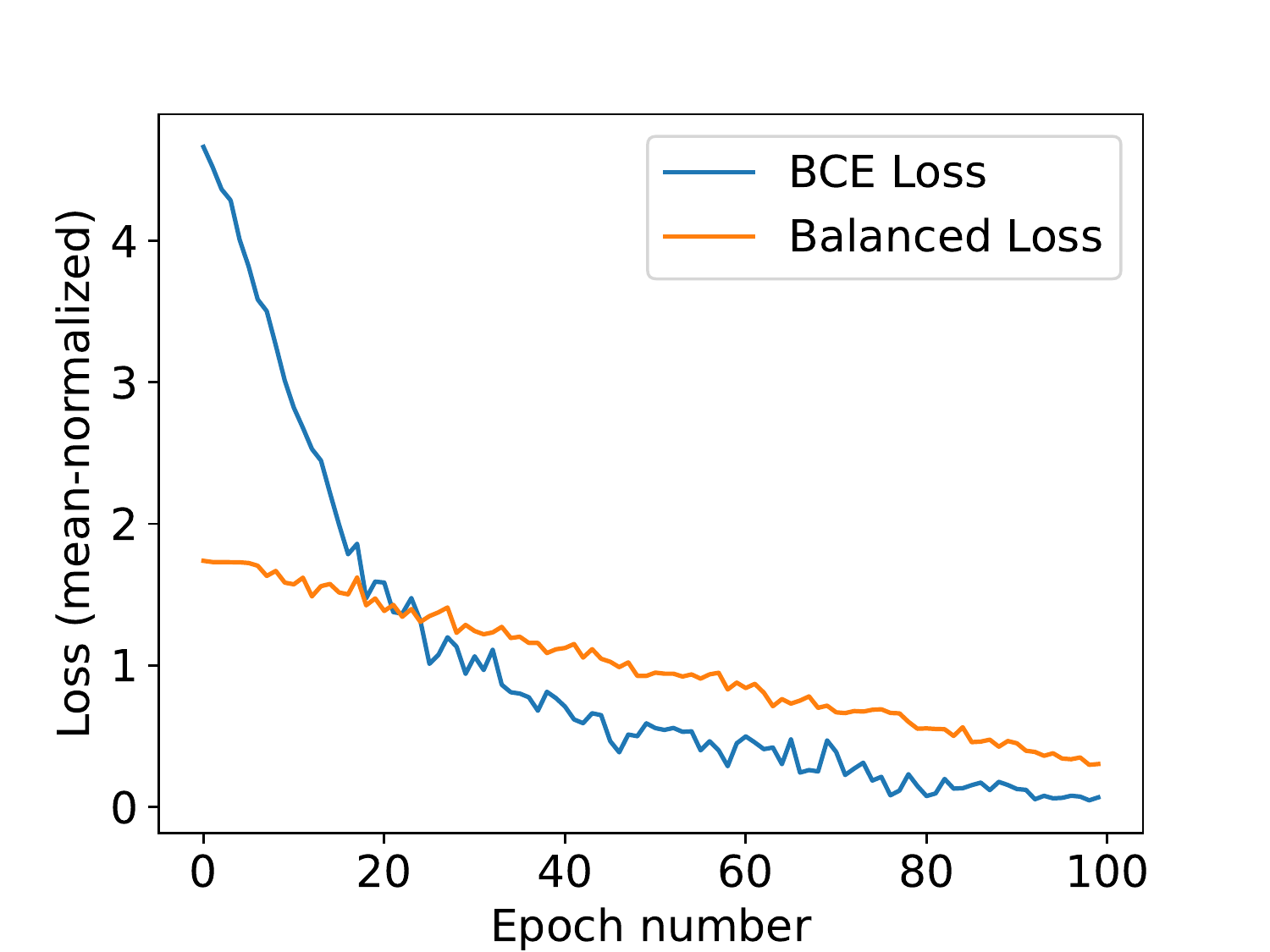}
    \caption{training\label{fig:comparison-train}}
  \end{subfigure}
  \hfill
  \begin{subfigure}[h]{0.45\linewidth}
    \includegraphics[width=\textwidth]{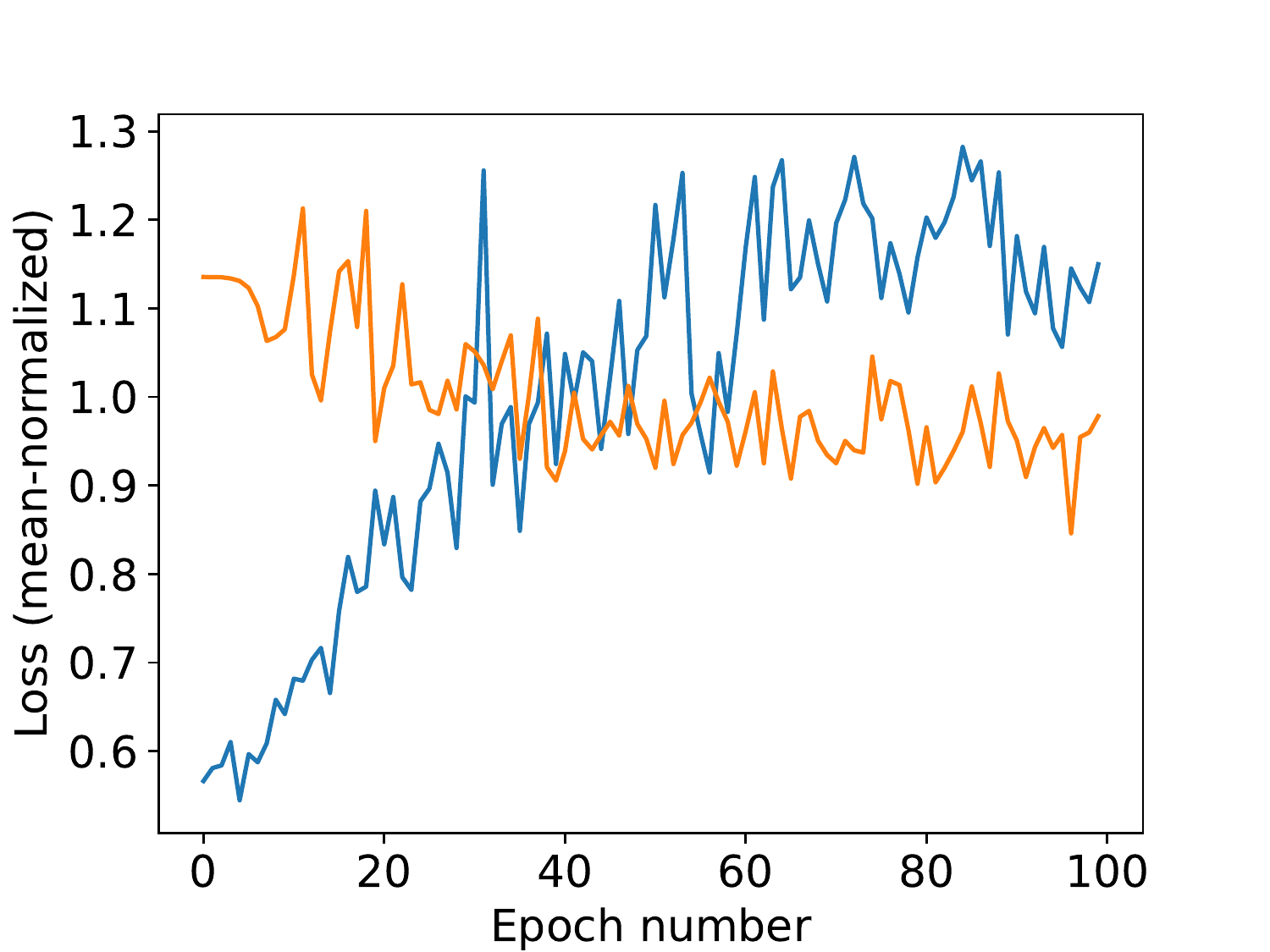}
    \caption{validation\label{fig:comparison-valid}}
  \end{subfigure}
  \caption{BCE loss and the balanced loss on the training set (left panel) and validation set (right panel) plotted as a function of epochs.}
  \label{fig:comparison}
\end{figure}

\begin{figure}[h]
  \includegraphics[width=\textwidth]{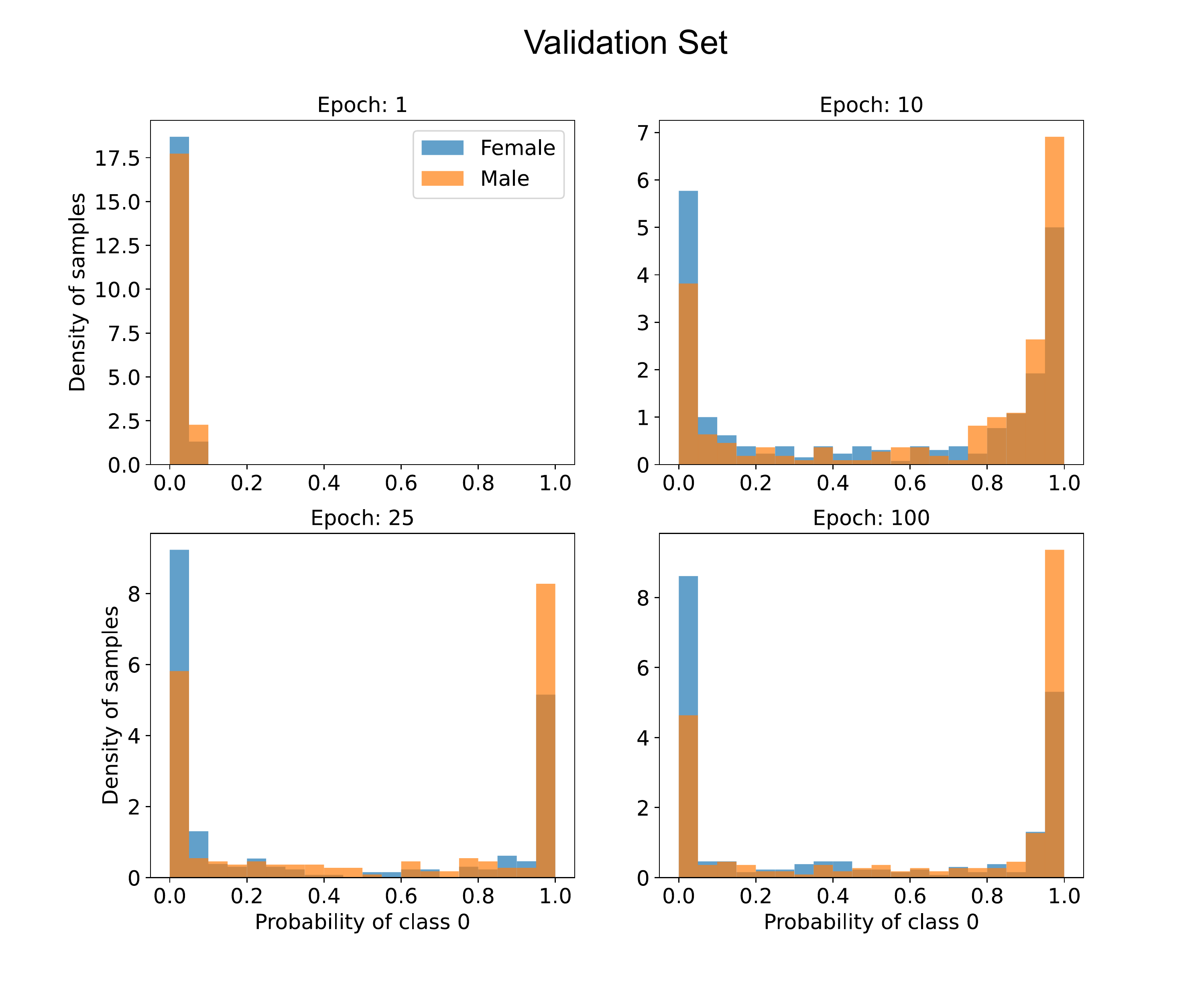}
  \caption{Histograms of model beliefs for validation data when trained with the
    balanced loss. Histograms are stratified by training epochs. Blue denotes
    female samples (class $0$), orange denotes male samples (class $1$). This
    graphic is a direct analogue of~\autoref{fig:odirN_validation_histograms}
    for the balanced loss. }
  \label{fig:balanced-loss-valid-hist}
\end{figure}

To test this hypothesis, we trained a model on ODIR-N once with BCE loss and
once with a ``balanced'' cosine loss, as a comparison. When the true label is
one, the balanced loss is defined by
$\mathrm{loss}(p \mid y = 1) := 1 + \cos(\pi p)$, as represented
in~\autoref{fig:conditional-balanced-loss}; analogously,
$\mathrm{loss}(p \mid y = 0) := 1 - \cos(\pi p)$. The methodological details of
these supplementary experiments are identical to those of the main experiments
reported in~\nameref{sec:results-main}. The choice of the new loss function was
arbitrary in the sense that the only constraint be a balanced change in penalty
with respect to ``confidence'' for incorrect and correct decisions. In other
words, an increase in loss due to increased confidence on an incorrect
classification is bounded and comparable to a decrease in loss when incorrect
decisions are replaced with correct ones. 
As shown in~\autoref{fig:comparison}, validation loss decreased during training
with the new bounded loss, in contrast to training with BCE loss. This occurred
even though the model still learned confident incorrect classifications on the
validation data (see~\autoref{fig:balanced-loss-valid-hist}). In conclusion,
this experiment illustrates that the increasing pattern of validation loss
observed during training can be remedied by replacing BCE loss with another.

\section*{S1 Bootstrap method}
\label{S1_bootstrap_method}

For a more thorough background on ideas related to and making use of
bootstrapping, refer to~\cite[$\S\,6.2$]{efron1994introduction}, from which this
section draws its main ideas.

Suppose that $x = (x_{1}, \ldots, x_{n})$ is a random sample, with
$x_{i} \sim \mathcal{F}, i \in [n]$, where $\mathcal{F}$ is a
distribution. Further suppose we wish to estimate a parameter of interest
$\theta = t(\mathcal{F})$ using the sample $x$, where $t$ is some mapping
yielding the ``true'' quantity $\theta$ corresponding to the distribution
$\mathcal{F}$. In order to estimate $\theta$, suppose we calculate
$\hat \theta := s(x)$ for a mapping $s$ that takes the random sample $x$ to a
quantity $\hat \theta$. For example, one might have
$s(x) := t(\hat{\mathcal{F}})$ where $\hat{\mathcal{F}}$ is the empirical
distribution constructed from $x$:
\begin{align*}
  \hat{\mathcal{F}} := \sum_{i\in [n]} \delta_{x_{i}}.
\end{align*}

In the present setting, we might have $x \in (\{0, 1\} \times [0, 1])^{n}$ with
$x_{i}, i \in [n]$ being the label for image $i$ and the classifier's
probability that image $i$ is class ``male''. Thus, $t$ gives the
$\AUC$ for the true distribution of images $\mathcal{F}$, and we
approximate the true distribution by computing $s(x) := \AUC(x)$ where
$\AUC$ is given in~\eqref{eq:def-auroc}.

Define a bootstrap sample of size $n$ by
$x^{*} := (x^{*}_{1}, \ldots, x^{*}_{n})$ where
$x^{*}_{i}\sim \hat{\mathcal{F}}$ are sampled with replacement from
$\hat{\mathcal{F}}$. $x^{*}$ is a randomized, resampled version of $x$. The idea
of bootstrapping is to evaluate the quality of $\hat \theta$ by analyzing the
distribution of $\hat \theta^{*} := s(x^{*})$.

It is worth noting that the bootstrap estimate of the standard error of a
statistic $\hat \theta$ is a plug-in estimate. Namely, the ideal bootstrap
estimate of the standard error of $\hat \theta$ is
$\mathrm{se}_{\hat{\mathcal{F}}}(\hat \theta^{*})$. As a recipe for the
so-called non-parametric bootstrap for the standard error of a statistic
$\hat \theta$:
\begin{enumerate}
\item Generate $(x^{* (b)})_{b=1}^{B}$ where each $x^{* (b)}$ has $n$ samples
  drawn from $\hat{\mathcal{F}}$;
\item Compute $\hat \theta^{*}(b) := s(x^{* (b)}), b \in [B]$;
\item Estimate $\mathrm{se}_{\mathcal{F}}(\hat \theta) \approx \widehat{\mathrm{se}}_{B}$
\end{enumerate}

\textbf{Note:} in this work, we do not estimate the standard error, but rather
quantiles. See~\citet{efron1994introduction, ghosh1984note, babu1986note} for
more on bootstrapping and bootstrapping quantiles, in particular.


\clearpage
\section*{S1 Parameter values}
\label{S1_parameter_values}

\subsection*{Runs D1--5}
\label{sec:pv-dovs-i}

Below, we list the parameter values used for the sex classification task using
the DOVS-i database. The same parameter settings were used for runs D1 through
D5.

\begin{adjustwidth}{-.5in}{.25in}
  \begin{minipage}[h]{\linewidth}
    \begin{multicols}{2}
      \begin{description}
      \item[Optimizer] \textbf{::}
        \begin{description}
        \item[method:] SGD 
        \item[learning rate:] $10^{-3}$ 
        \item[batch size:] $16$ 
        \item[weight decay:] $10^{-3}$ 
        \item[acceleration:] Nesterov 
        \item[momentum:] $0.9$
        \item[annealing:] $\mathtt{ExponentialLR}(0.99)$  
        \end{description}
      \item[Criterion] \textbf{::}
        \begin{description}
        \item[Max epochs:] $1000$
        \item[early stopping:] validation AUC
        \item[loss:] binary cross-entropy
        \item[class weights (F, M):] $(0.98, 1.02)$
        \end{description}
      \item[Network] \textbf{::}
        \begin{description}
        \item[base network:] ResNet-152
        \item[hidden layers:] $2048$
        \item[dropout probability:] $0.5$
        \end{description}
      \item[Transforms] \textbf{::}
        \begin{description}
        \item[Train] \textbf{::} 
\begin{verbatim}
ColorJitter(
    brightness=[0.95, 1.05], 
    contrast=[0.95, 1.05], 
    saturation=[0.95, 1.05], 
    hue=[-0.05, 0.05]
),
RandomHorizontalFlip(p=0.5),
RandomVerticalFlip(p=0.5),
Normalize(
    mean=[0.485, 0.456, 0.406], 
    std=[0.229, 0.224, 0.225]
)
\end{verbatim}
        \item[Val \& Test] \textbf{::} 
\begin{verbatim}
Normalize(
    mean=[0.485, 0.456, 0.406], 
    std=[0.229, 0.224, 0.225]
)
\end{verbatim}
          \vphantom{A}
        \end{description}
      \end{description}
    \end{multicols}
  \end{minipage}
\end{adjustwidth}

\clearpage
\subsection*{Runs N1--6}
\label{sec:pv-odir-n}

Below are the parameter values used in the sex classification task trained on
the ODIR-N database. The same parameter settings were used for runs N1 through
N6, except for the omission of channel normalization in the image preprocessing
transform for runs N1 through N3. In runs N1 through N3, the preprocessed images
were not normalized (using \texttt{transforms.Normalize}) before being fed into
the network. This is reflected in the description below. In runs N4 through N6,
normalization of the channels was applied (using \texttt{transforms.Normalize})
with the same mean and variance as that for DOVS-i and ODIR-C (\emph{cf.}
below). When the channel normalization was used, it was applied to training,
validation and test images alike, as the final transform before images were fed
into the network.

\begin{adjustwidth}{-.5in}{.25in}
  \begin{minipage}[h]{\linewidth}
    \begin{multicols}{2}
      \begin{description}
      \item[Optimizer] \textbf{::}
        \begin{description}
        \item[method:] SGD 
        \item[learning rate:] $10^{-3}$ 
        \item[batch size:] $16$ 
        \item[weight decay:] $10^{-3}$ 
        \item[acceleration:] Nesterov 
        \item[momentum:] $0.9$ 
        \item[annealing:] $\mathtt{ExponentialLR}(0.99)$  
        \end{description}
      \item[Criterion] \textbf{::}
        \begin{description}
        \item[Max epochs:] $1000$
        \item[early stopping:] validation AUC
        \item[loss:] binary cross-entropy
        \item[class weights (F, M):] $(0.91, 1.11)$
        \end{description}
      \item[Network] \textbf{::}
        \begin{description}
        \item[base network:] ResNet-152
        \item[hidden layers:] $1024$
        \item[dropout probability:] $0.3$
        \end{description}
      \item[Transforms] \textbf{::}
        \begin{description}
        \item[Train] \textbf{::} 
\begin{verbatim}
ColorJitter(
    brightness=[0.95, 1.05], 
    contrast=[0.95, 1.05], 
    saturation=[0.95, 1.05], 
    hue=[-0.05, 0.05]
),
RandomHorizontalFlip(p=0.5),
RandomVerticalFlip(p=0.5)
\end{verbatim}
        \end{description}
      \end{description}
    \end{multicols}
  \end{minipage}
\end{adjustwidth}

\clearpage
\subsection*{Runs C1--6}
\label{sec:pv-odir-c}

These are the parameter values for the sex classification task trained using the
ODIR-C database. The same parameters were used for runs C1 through C6.

\begin{adjustwidth}{-.5in}{.25in}
  \begin{minipage}[h]{\linewidth}
    \begin{multicols}{2}
      \begin{description}
      \item[Optimizer] \textbf{::}
        \begin{description}
        \item[method:] SGD 
        \item[learning rate:] $10^{-3}$ 
        \item[batch size:] $16$ 
        \item[weight decay:] $10^{-3}$ 
        \item[acceleration:] Nesterov 
        \item[momentum:] $0.9$ 
        \item[annealing:] $\mathtt{ExponentialLR}(0.99)$  
        \end{description}
      \item[Criterion] \textbf{::}
        \begin{description}
        \item[Max epochs:] $1000$
        \item[early stopping:] validation AUC
        \item[loss:] binary cross-entropy
        \item[class weights (F, M):] $(0.92, 1.10)$
        \end{description}
      \item[Network] \textbf{::}
        \begin{description}
        \item[base network:] ResNet-152
        \item[hidden layers:] $2048$
        \item[dropout probability:] $0.5$
        \end{description}
      \item[Transforms] \textbf{::}
        \begin{description}
        \item[Train] \textbf{::} 
\begin{verbatim}
ColorJitter(
    brightness=[0.95, 1.05], 
    contrast=[0.95, 1.05], 
    saturation=[0.95, 1.05], 
    hue=[-0.05, 0.05]
),
RandomHorizontalFlip(p=0.5),
RandomVerticalFlip(p=0.5),
Normalize(
    mean=[0.485, 0.456, 0.406], 
    std=[0.229, 0.224, 0.225]
)
\end{verbatim}
        \item[Val \& Test] \textbf{::} 
\begin{verbatim}
Normalize(
    mean=[0.485, 0.456, 0.406], 
    std=[0.229, 0.224, 0.225]
)
\end{verbatim}
          \vphantom{A}
        \end{description}
      \end{description}
    \end{multicols}
  \end{minipage}
\end{adjustwidth}

\clearpage
\subsection*{Runs E1--20}
\label{sec:pv-ensemble-dovs-ii}

In this section we include the parameter values used in the sex classification
task that used an ensemble of $10$ ResNet-152 models trained using the DOVS-ii
database. The same parameters were used for runs E1 through E20.

\begin{adjustwidth}{-.5in}{.25in}
  \begin{minipage}[h]{\linewidth}
    \begin{multicols}{2}
      \begin{description}
      \item[Optimizer] \textbf{::}
        \begin{description}
        \item[method:] SGD 
        \item[learning rate:] $10^{-3}$ 
        \item[batch size:] $16$ 
        \item[weight decay:] $10^{-3}$ 
        \item[acceleration:] Nesterov 
        \item[momentum:] $0.9$ 
        \item[annealing:] $\mathtt{ExponentialLR}(0.99)$  
        \end{description}
      \item[Criterion] \textbf{::}
        \begin{description}
        \item[Max epochs:] $1000$
        \item[early stopping:] validation AUC
        \item[loss:] binary cross-entropy
        \item[class weights (F, M):] $(0.98, 1.02)$
        \end{description}
      \item[Network] \textbf{::}
        \begin{description}
        \item[base network:] ResNet-152
        \item[hidden layers:] $2048$
        \item[dropout probability:] $0.5$
        \end{description}
      \item[Transforms] \textbf{::}
        \begin{description}
        \item[Train] \textbf{::} 
\begin{verbatim}
ColorJitter(
    brightness=[0.95, 1.05], 
    contrast=[0.95, 1.05], 
    saturation=[0.95, 1.05], 
    hue=[-0.05, 0.05]
),
RandomHorizontalFlip(p=0.5),
RandomVerticalFlip(p=0.5),
Normalize(
    mean=[0.485, 0.456, 0.406], 
    std=[0.229, 0.224, 0.225]
)
\end{verbatim}
        \item[Val \& Test] \textbf{::} 
\begin{verbatim}
Normalize(
    mean=[0.485, 0.456, 0.406], 
    std=[0.229, 0.224, 0.225]
)
\end{verbatim}
          \vphantom{A}
        \end{description}
      \end{description}
    \end{multicols}
  \end{minipage}
\end{adjustwidth}


\section*{S1 Data statistics}
\label{S1_data_statistics}

\subsection*{DOVS-i \& DOVS-ii}
\label{sec:dset-stats-dovs}

The original image sizes were $2392 \times 2048$. The images were resized to
$224 \times 224$, an image size that has been used for other image
classification tasks~\cite{he2016deep,howard2017mobilenets}. To perform the
resizing, the images were first resized to the nearest power of $2$ using a
discrete Haar wavelet transform~\cite{lee19_pywav}, and then resized to
$224 \times 224$ using bilinear intepolation \emph{via} a standard Python
package for image manipulation tasks~\cite{clark2015pillow}.

We note here that the DOVS-ii database is a superset of the DOVS-i database ---
a result of an additional round of data retrieval from the VGH
servers. Properties of and preprocessing for images in DOVS-i is the same as
that for DOVS-ii.

\begin{figure}[h]
  \centering
  \begin{flushleft}
    \textbf{DOVS-i}
  \end{flushleft}
  \includegraphics[width=\textwidth]{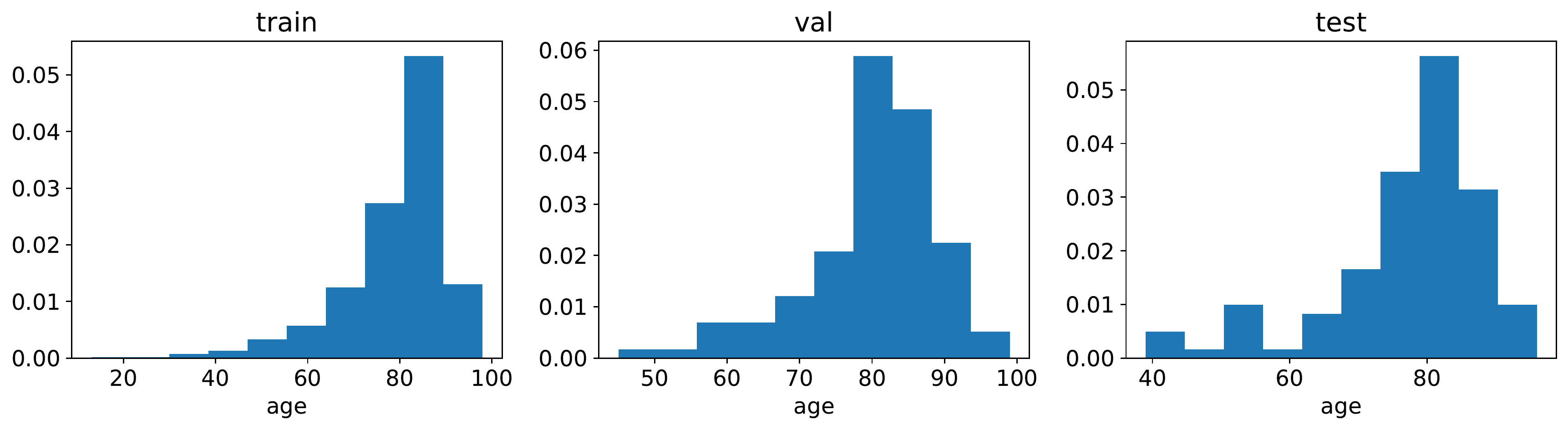}
  \begin{flushleft}
    \textbf{DOVS-ii}
  \end{flushleft}
  \includegraphics[width=\textwidth]{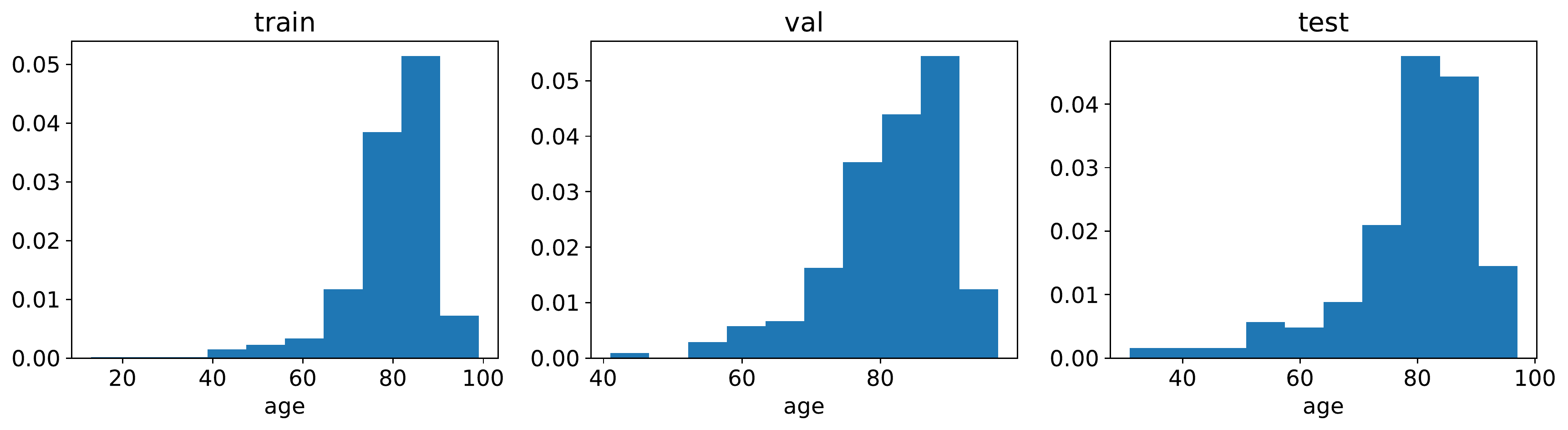}
  
  \caption{DOVS-i and DOVS-ii patient age statistics: phase-stratified
    density-style histograms representing the patient age distributions for each
    dataset.}
  \label{fig:patient-stats}
\end{figure}

Patient statistics are summarized in~\autoref{fig:patient-stats}
and~\autoref{tab:patient-stats}. Histograms for age data appear
in~\autoref{fig:patient-stats} for DOVS-i (top) and DOVS-ii
(bottom). Descriptive statistics stratified by patient age and sex appear
in~\autoref{tab:patient-stats-dovs-i} for DOVS-i
and~\autoref{tab:patient-stats-dovs-ii} for DOVS-ii.

\begin{table}[h!]
  \centering
    \begin{subtable}{\linewidth}
      \begin{tabular}{llllllllll}
        \toprule
        {} & \multicolumn{3}{l}{train} & \multicolumn{3}{l}{val} & \multicolumn{3}{l}{test} \\
        {} &      F &      M &    $\cup$ &      F &      M &    $\cup$ &      F &      M &    $\cup$ \\
        \midrule
        min  &  $24$ &  $13$ &  $13$ &  $55$ &  $45$ &  $45$ &  $53$ &  $39$ &  $39$ \\
        max  &  $98$ &  $97$ &  $98$ &  $96$ &  $99$ &  $99$ &  $95$ &  $96$ &  $96$ \\
        mean &  $79.59$ &  $78.06$ &  $78.84$ &  $80.34$ &  $79.73$ &  $80.05$ &  $79.44$ &  $74.54$ &  $77.04$ \\
        st.dev. &  $11.76$ &  $10.96$ &  $11.40$ &  $~\,8.05$ &  $10.31$ &  $~\,9.18$ &  $~\,8.88$ &  $13.31$ &  $11.51$ \\
        \bottomrule
      \end{tabular}
      \caption{DOVS-i age statistics.\label{tab:patient-stats-dovs-i}}
    \end{subtable}
    \vskip6pt
    
    \begin{subtable}{\linewidth}
      \begin{tabular}{llllllllll}
        \toprule
        {} & \multicolumn{3}{l}{train} & \multicolumn{3}{l}{val} & \multicolumn{3}{l}{test} \\
        {} &      F &      M &    $\cup$ &      F &      M &    $\cup$ &      F &      M &    $\cup$ \\
        \midrule
        $\min$ &  $24$ &  $13$ &  $13$ &  $55$ &  $41$ &  $41$ &  $31$ &  $37$ &  $31$ \\
        $\max$ &  $98$ &  $99$ &  $99$ &  $97$ &  $95$ &  $97$ &  $97$ &  $97$ &  $97$ \\
        mean &  $80.17$ &  $79.18$ &  $79.68$ &  $82.93$ &  $79.36$ &  $81.17$ &  $79.58$ &  $78.60$ &  $79.10$ \\
        st.dev. &  $~\,9.60$ &  $~\,9.65$ &  $~\,9.63$ &  $~\,8.17$ &  $~\,9.73$ &  $~\,9.14$ &  $11.60$ &  $11.07$ &  $11.34$ \\
        \bottomrule
      \end{tabular}
      \caption{DOVS-ii age statistics.\label{tab:patient-stats-dovs-ii}}
    \end{subtable}
  \vskip6pt

  \begin{subtable}{\linewidth}    
    \begin{tabular}{llllllllll}
      \toprule
      & \multicolumn{3}{l}{train} & \multicolumn{3}{l}{val} & \multicolumn{3}{l}{test} \\
              & F    & M    & $\cup$ & F    & M    & $\cup$ & F    & M    & $\cup$ \\
      \midrule
      min     & $~\,1$ & $17$   & $~\,1$ & $28$   & $25$   & $25$   & $22$   & $15$   & $15$   \\
      max     & $89$   & $89$   & $89$   & $85$   & $87$   & $87$   & $88$   & $91$   & $91$   \\
      mean    & $58.6$ & $56.4$ & $57.4$ & $59.3$ & $56.1$ & $57.6$ & $58.8$ & $55.7$ & $57.1$ \\
      st.dev. & $11.5$ & $11.0$ & $11.3$ & $10.6$ & $11.6$ & $11.3$ & $11.7$ & $11.5$ & $11.6$ \\
      \bottomrule
    \end{tabular}
    \caption{ODIR-N age statistics.\label{tab:odir-n-age-stats}}
  \end{subtable}
  \vskip6pt

  \begin{subtable}{\linewidth}
    \begin{tabular}{llllllllll}
      \toprule
      & \multicolumn{3}{l}{train} & \multicolumn{3}{l}{val} & \multicolumn{3}{l}{test} \\
      & F & M & $\cup$ & F & M & $\cup$ & F & M & $\cup$ \\
      \midrule
      min     & $~\,1$ & $15$   & $~\,1$ & $~\,1$ & $17$   & $~\,1$ & $26$   & $31$   & $26$   \\
      max     & $87$   & $87$   & $87$   & $87$   & $82$   & $87$   & $82$   & $89$   & $89$   \\
      mean    & $58.1$ & $56.0$ & $57.0$ & $56.4$ & $54.2$ & $55.2$ & $58.9$ & $55.4$ & $57.0$ \\
      st.dev. & $10.9$ & $11.1$ & $11.0$ & $12.7$ & $10.5$ & $11.6$ & $10.2$ & $10.9$ & $10.7$ \\
      \bottomrule
    \end{tabular}
    \caption{ODIR-C age statistics.\label{tab:odir-c-age-stats}}
  \end{subtable}
  \caption{Patient age statistics: phase- and sex-stratified summary statistics
    for patient age within each dataset. The ``F'' columns correspond to
    statistics for female patients; ``M'', male patients; $\cup$, no
    stratification by sex. \label{tab:patient-stats}}
\end{table}

\subsection*{ODIR}
\label{sec:dset-stats-odir}

The ODIR database was curated by hosts of an online competition for machine
learning on ocular disease~\cite{odir-database}. This database contains 7000
annotated images from 3500 patients --- a left/right pair of images for each
patient in the database.

\subsection*{ODIR-N}
\label{sec:dset-stats-odir-n}

The ODIR-N variant of the ODIR database was curated by us. It was obtained by
subsampling the ODIR database, retaining only those eye images that were
``normal'' (\ie no adverse health condition annotation was present). After
subsampling of the database, the images were cropped to a bounding box about the
fundus image and downsampled to a size of $224 \times 224$ as described in
\nameref{sec:dset-stats-dovs}. Finally, the colour channels of the images were
adjusted using CLAHE~\cite{albumentations}. We coin as ODIR-N the resulting
dataset of $3098$ images from $1959$ individuals. The patients in this database
were randomly partitioned into three datasets for training, validation and test
sets, comprising $70\%$, $15\%$ and $15\%$ of the patients, respectively
(\emph{cf.}~\autoref{tab:odir-n-dset-stats}
). See \autoref{tab:odir-n-age-stats} for patient age statistics stratified by
phase, with and without stratification by patient sex.

\subsection*{ODIR-C}
\label{sec:dset-stats-odir-c}

The ODIR-C database is a subset of the ODIR-N database, obtained by admitting or
omitting images through a quality vetting procedure. The train/val/test
partitions of the ODIR-C images are different than those of the ODIR-N
images. For example, it is very likely that there exist two images in the train
set of ODIR-N such that one resides in the train set of ODIR-C and the other in
the test set of ODIR-C. However, the proportions in each partition are the same
(\emph{cf}.~\autoref{tab:odir-c-dset-stats}). See \autoref{tab:odir-c-age-stats}
for patient age statistics stratified by phase, with and without stratification
by patient sex.

We now describe the vetting procedure, which effectively follows a combined
ruleset of those used by~\citet{gulshan2016development}
and~\citet{ting2017development}. Specifically, images from ODIR-N are admitted/omitted on the basis of the following items. 

\begin{enumerate}
\item Illumination: Is the image too dark, or too light? Are there dark areas or washed-out areas?
\item Image field definition: Does the primary field include the entire optic nerve head and macula? 
\item Artifacts: Is the image sufficiently free of artifacts (\eg dust spots,
  arc defects, and eyelash images)?
\item Validity: Is the image a valid ``retinal image''? \label{it:valid}
\item Compositeness: Is the image acquired from a single capture event with normal angle of view which provides a 30 to 50 degree image or was it
  composited? \label{it:composite}
\end{enumerate}
In the case of~\ref{it:valid}, a retinal image may be \emph{invalid}, for example, if there
is media opacity (such as cataracts). This exclusion is consistent
with~\citet{ting2017development}. Regarding~\ref{it:composite}, composite images were excluded
from ODIR-C.


\end{document}